\documentclass[lettersize,journal]{IEEEtran}
\usepackage{amsmath,amsfonts}
\usepackage{algorithm}
\usepackage{array}
\usepackage[caption=false,font=normalsize,labelfont=sf,textfont=sf]{subfig}
\usepackage{textcomp}
\usepackage{stfloats}
\usepackage{url}
\usepackage{verbatim}
\usepackage{graphicx}
\usepackage{cite}
\usepackage{amssymb}
\usepackage{multirow}%
\usepackage{amsthm}%
\usepackage{mathrsfs}%
\usepackage{xcolor}%
\usepackage{manyfoot}%
\usepackage{booktabs}%
\usepackage{colortbl}
\usepackage{algpseudocode}%
\usepackage{bbm}
\usepackage{wrapfig}
\usepackage{subcaption}
\usepackage{dsfont}
\usepackage{arydshln}
\usepackage{booktabs}
\newcommand{\ie}{\emph{i.e.}}
\newcommand{\eg}{\emph{e.g.}}
\newcommand{\darkgreen}[1]{\textcolor[rgb]{0.00,0.70,0.00}{#1}}
\newcommand{\darkred}[1]{\textcolor[rgb]{0.70,0.00,0.00}{#1}}

\begin{document}

\title{GoMatching++: Parameter- and Data-Efficient Arbitrary-Shaped Video Text Spotting and Benchmarking}

\author{Haibin He, Jing Zhang~\IEEEmembership{Senior~Member,~IEEE,}, Maoyuan Ye, Juhua Liu~\IEEEmembership{Member,~IEEE,}, Bo Du~\IEEEmembership{Senior~Member,~IEEE,}, Dacheng Tao~\IEEEmembership{Fellow,~IEEE}

\thanks{This work was supported in part by the National Key Research and Development Program of China under Grant 2023YFC2705700, in part by the National Natural Science Foundation of China under Grants U23B2048 and 62225113, in part by the Innovative Research Group Project of Hubei Province under Grant 2024AFA017, and in part by the Science and Technology Major Project of Hubei Province under Grants 2024BAB046 and 2025BCB026. (\textit{Corresponding Authors: Jing Zhang, Juhua Liu})}

\thanks{Haibin He, Jing Zhang, Maoyuan Ye, Juhua Liu, Bo Du are with the School of Computer Science, National Engineering Research Center for Multimedia Software, and Institute of Artificial Intelligence, Wuhan University, China (e-mail: haibinhe@whu.edu.cn; jingzhang.cv@gmail.com; yemaoyuan@whu.edu.cn; liujuhua@whu.edu.cn; dubo@whu.edu.cn)}

\thanks{Dacheng Tao are with the College of Computing \& Data Science, Nanyang Technological University, Singapore (e-mail: dacheng.tao@gmail.com)}

}
\markboth{Journal of \LaTeX\ Class Files,~Vol.~14, No.~8, August~2021}%
{Shell \MakeLowercase{\textit{et al.}}: A Sample Article Using IEEEtran.cls for IEEE Journals}

\maketitle

\begin{abstract}
Video text spotting (VTS) extends image text spotting (ITS) by adding text tracking, significantly increasing task complexity. Despite progress in VTS, existing methods still fall short of the performance seen in ITS. This paper identifies a key limitation in current video text spotters: limited recognition capability, even after extensive end-to-end training. To address this, we propose \textit{GoMatching++}, a parameter- and data-efficient method that transforms an off-the-shelf image text spotter into a video specialist. The core idea lies in freezing the image text spotter and introducing a lightweight, trainable tracker, which can be optimized efficiently with minimal training data. Our approach includes two key components: (1) a rescoring mechanism to bridge the domain gap between image and video data, and (2) the LST-Matcher, which enhances the frozen image text spotter's ability to handle video text. We explore various architectures for LST-Matcher to ensure efficiency in both parameters and training data. As a result, GoMatching++ sets new performance records on challenging benchmarks such as ICDAR15-video, DSText, and BOVText, while significantly reducing training costs. To address the lack of curved text datasets in VTS, we introduce \textit{ArTVideo}, a new benchmark featuring over 30\% curved text with detailed annotations. We also provide a comprehensive statistical analysis and experimental results for ArTVideo. We believe that GoMatching++ and the ArTVideo benchmark will drive future advancements in video text spotting. The source code, models and dataset are publicly available at \url{https://github.com/Hxyz-123/GoMatching}.
\end{abstract}


\section{Introduction}\label{intro}
\begin{figure*}[t]
\centering
  \subfloat[]{\includegraphics[width=0.47\textwidth]{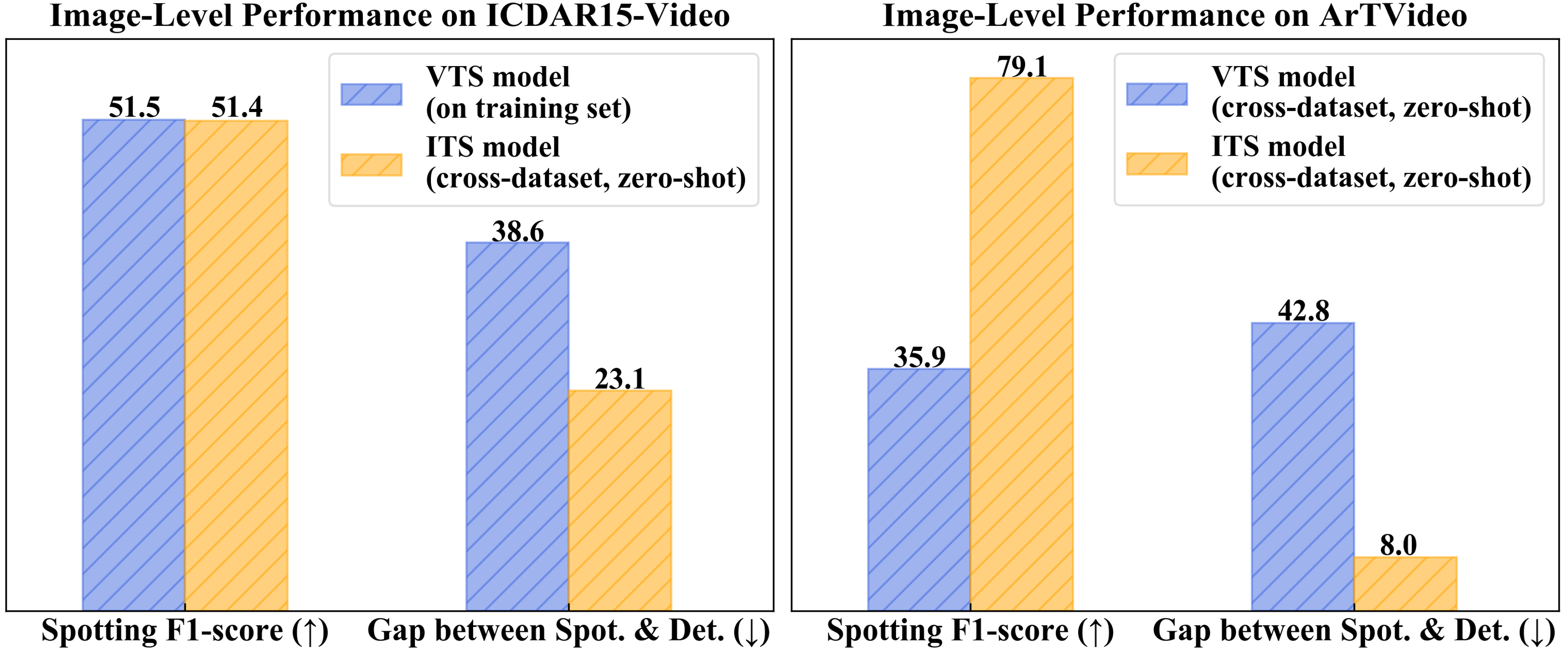}}
  \hfill
  \subfloat[]{\includegraphics[width=0.52\textwidth]{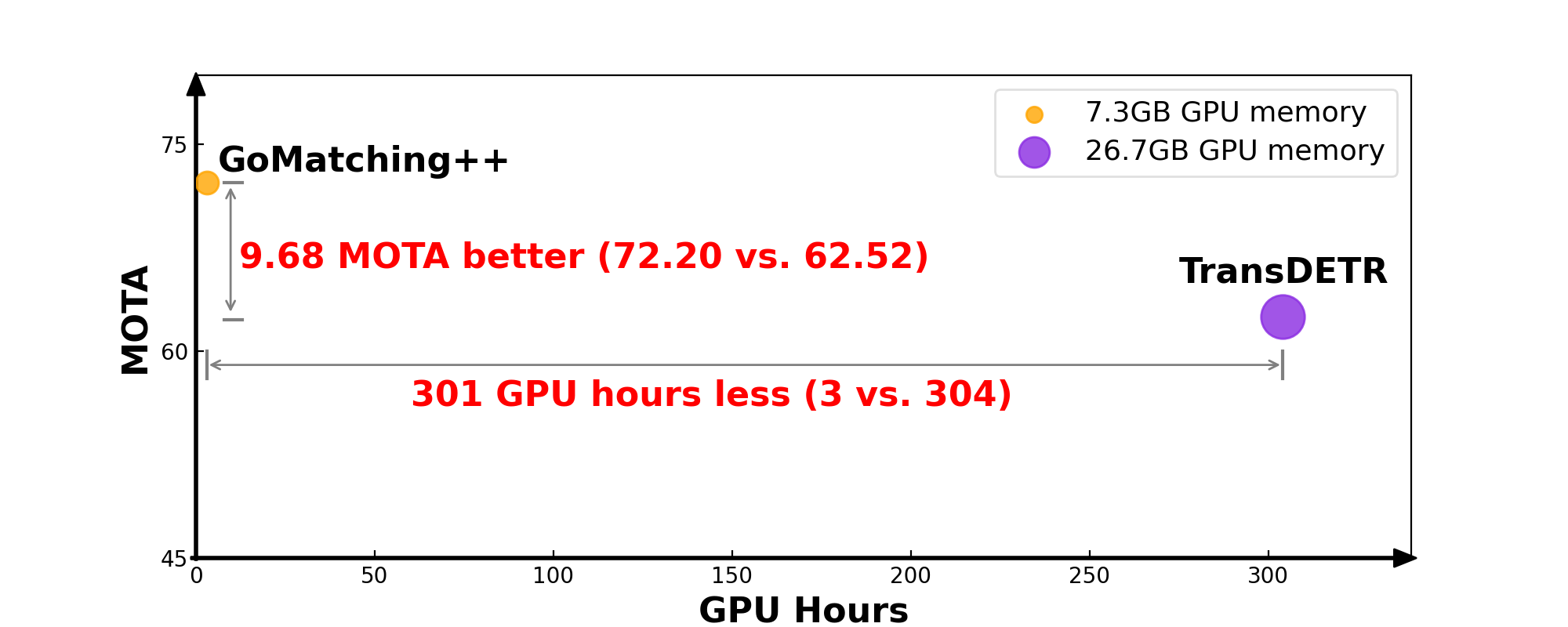}}
  \caption{(a) `Gap between Spot. \& Det.': the gap between spotting and detection F1-score. As the spotting task involves recognizing the results of the detection process, the detection score is indeed the upper bound of spotting performance. The larger the gap, the poorer the recognition ability. Compared to the ITS model (Deepsolo~\cite{ye2023deepsolo}), the VTS model (TransDETR~\cite{wu2024end}) presents unsatisfactory image-level text spotting F1-scores, which lag far behind its detection performance, especially on ArTVideo with curved text. It indicates recognition capability is a main bottleneck in the VTS model.
  (b) GoMatching++ outperforms TransDETR by over 9.68 MOTA on ICDAR15-video while saving 301 training GPU hours and 19.4GB memory. }
 \label{fig:1}
\end{figure*}

With the burgeoning growth of visual information technology, text spotting has emerged as an effective tool for various applications, such as video understanding~\cite{srivastava2015unsupervised, zhao2022towards, tom2023reading}, video retrieval~\cite{dong2021dual} and autonomous driving~\cite{zhang2021character,zablocki2022explainability}. Recently, numerous image text spotting (ITS) methods~\cite{long2021scene, feng2021residual, liu2023spts,zhang2022text,ye2023deepsolo,huang2023estextspotter} that simultaneously tackle text detection and recognition, have attained extraordinary accomplishment. In the video realm, video text spotting (VTS) introduces an additional tracking task, intensifying the challenge to a new level. Although VTS methods~\cite{wang2017end,cheng2019you,cheng2020free,wu2021bilingual,wu2024bilingual,wu2024end} have made significant progress, a substantial discrepancy persists when compared to ITS. We observe that the text recognition proficiency of VTS models is far inferior to ITS models, even when extensively trained end-to-end on video data.

To investigate this disparity, we compare the state-of-the-art (SOTA) VTS model TransDETR~\cite{wu2024end} and ITS model Deepsolo~\cite{ye2023deepsolo} for image-level text spotting performance on ICDAR15-video~\cite{Karatzas2015ICDAR15} and our ArTVideo (\textit{i.e.}, \textbf{Ar}bitrary-shaped \textbf{T}ext in \textbf{Video}) (Sec.\ref{art}), which comprises over 30\% curved text. As illustrated in Fig.~\ref{fig:1}(a), even when evaluating the image-level spotting performance on the VTS model's training set, the F1-score of TransDETR is merely comparable to the zero-shot performance of Deepsolo. 
The performance of the VTS model on ArTVideo is much worse. Moreover, the significant disparity between the spotting and detection-only performance of the VTS model highlights that its recognition capability constitutes the main bottleneck.
We attribute this discrepancy to two key aspects: 1) the model architecture and 2) the training data. First, in terms of model architecture, some studies~\cite{zhang2023motrv2,yu2023motrv3} have indicated that there exists optimization conflict in detection and association during the end-to-end training of MOTR~\cite{zeng2022motr}. We hold that TransDETR~\cite{wu2024end}, which integrates text recognition into a MOTR-based design, is likely to incur more serious conflicts, alongside requiring extensive training resources and data in an end-to-end framework. 
On the other hand, ITS studies~\cite{ye2023deepsolo,huang2023estextspotter} have presented the advantages of employing advanced query formulation for text spotting in DETR frameworks~\cite{carion2020end,zhu2020deformable}. In contrast, existing Transformer-based VTS models still rely on Region of Interest (RoI) components or simply cropping detected text regions for recognition.
Second, with respect to the training data, most text instances in current video text dataset~\cite{Karatzas2015ICDAR15,wu2021bilingual,Wu2023DSText} are either horizontal or oriented, and the bounding box annotations are limited to quadrilaterals. This constrains both the data diversity and the recognition performance, particularly for curved text. 
Hence, \textit{leveraging model and data knowledge from ITS presents considerable value for VTS}.
To achieve this, an intuitive idea is to exploit an off-the-shelf SOTA image text spotter. However, the utilization in a naive manner gives rise to two fundamental problems: 1) video frames often feature a higher frequency of small and blurry text instances caused by camera motion and varying distances. This domain gap leads to low detection confidence and consequently a relatively low recall for the image text spotter, and 2) native image text spotters lack the ability to track the text across frames.

To this end, we propose GoMatching++, a simple and effective baseline that adheres to the tracking-by-detection paradigm. Specifically, building on a frozen ITS model (\eg DeepSolo~\cite{ye2023deepsolo}), we introduced the rescoring mechanism along with a tailored tracker, LST-Matcher. The rescoring mechanism involves an additional lightweight, trainable text classification head called rescoring head via efficient tuning on video data and recalibrating confidence scores for detected instances. Each instance's final score is computed through a fusion operation between the original score from the image text spotter and the recalibrated score produced by the rescoring head. This mechanism preserve the ITS model's inherent knowledge while alleviating performance degradation arising from the image-video domain gap. Subsequently, the identified text instances are sent to LST-Matcher for association. 
LST-Matcher, consisting of a shared feedforward network (FFN) and dual parallel Transformers with single-layer encoder-decoder architectures, can effectively harness both long- and short-term information, endowing it a high-capable tracker. Since both the rescoring head and LST-Matcher are lightweight and gradient-decoupled, GoMatching++ significantly reduces training resources and data requirements while avoiding optimization conflicts. 

Furthermore, we explore more concise and effective architectures for LST-Matcher with only 30\% of the trainable parameter compared to TransDETR.
These designs conserve considerable trainable parameters and facilitate better tracker training with less data, resulting in a highly parameter- and data-efficient VTS model. As a result, our model significantly surpasses existing SOTA methods by a large margin with much lower training costs, as shown in Fig.~\ref{fig:1}(b). In addition to GoMatching++, we also establish a benchmark (ArTVideo) that contains 60 videos with over 30\% curved text to fill the gap in the research realm of video curved text spotting.

This work is an extensive version of our conference paper~\cite{he2024gomatching}, offering four additional contributions: (1) We \textbf{explore more concise architectural designs for GoMatching and propose GoMatching++}, a novel parameter- and data-efficient model. (2) We \textbf{expand ArTVideo from a simple test set to a comprehensive benchmark, including both training and test sets}, to mitigate the scarcity of curved text datasets in the VTS domain. (3) We \textbf{conduct extensive experiments to ablate and analyze the efficiency and effectiveness of GoMatching++}. (4) We \textbf{provide an in-depth statistical and experimental analysis of ArTVideo}, providing new insights into VTS.

The rest of this paper is organized as follows. In Sec.~\ref{works}, we briefly review the related work. In Sec.~\ref{med}, we introduce our proposed method in detail. Sec.~\ref{art} provides a comprehensive overview of the benchmark. Sec.~\ref{exp_open} and Sec.~\ref{exp_art} report and discuss our experimental results on existing open-source datasets and our ArtVideo, respectively. Lastly, we conclude our study in Sec.~\ref{conclusion}.

\section{Related Work}\label{works}
\subsection{Image Text Spotting}\label{works_its}
Early approaches~\cite{liao2020mask,wang2021pan++,liu2021abcnet} crafted RoI-based modules to bridge text detection and recognition. However, these methods ignored one vital issue, \ie, the synergy problem between the two tasks. 
To overcome this dilemma, recent Transformer-based methods~\cite{cao2021all,zhang2022text,liu2023spts,ye2023deepsolo,tong2024granularity} get rid of the fetters of RoI modules, and chase a better representation for the two tasks. For example, DETR-based TESTR~\cite{zhang2022text} uses two decoders for each task in parallel. In contrast, DeepSolo~\cite{ye2023deepsolo} proposes a unified and explicit query form for the two tasks, without harnessing dual decoders. However, the above methods cannot perform tracking in the video. 

\subsection{Video Text Spotting}\label{works_vts}
Compared to ITS, existing SOTA VTS methods still rely on RoI or simply cropping detected text regions for recognition. TransVTSpotter~\cite{wu2021bilingual} directly crops the detection regions from a transformer-based detector to perform recognition.
CoText~\cite{wu2024bilingual} adopts a text spotter with Masked-RoI and applies cosine similarity matching for tracking via a tracking head.
TransDETR~\cite{wu2024end} performs detection and tracking under the MOTR paradigm, using Rotated-RoI for the subsequent recognizer.
They pursue training all modules in an end-to-end manner. In comparison, we explore turning a RoI-free ITS model into a VTS one. We reveal the probability of freezing ITS part and focusing on tracking, thereby establishing a novel parameter- and data-efficient baseline while achieving SOTA performance.

\begin{figure*}[t]
  \centering
    \includegraphics[width=1.0\textwidth]{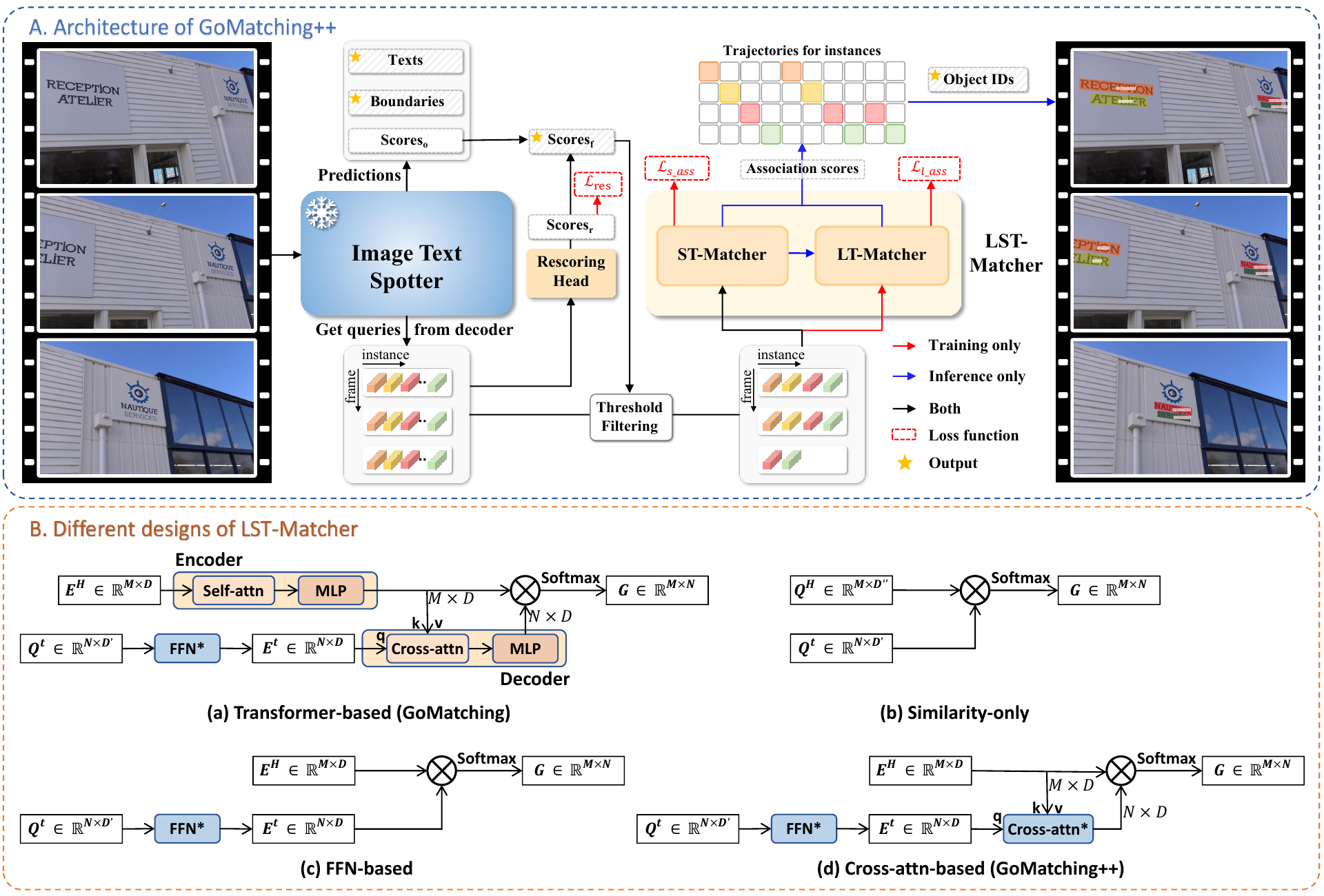}
    \caption{\textbf{A. The overall architecture of GoMatching++}. 
    The frozen image text spotter provides text spotting results for frames. The rescoring mechanism utilizes both instance scores from the image text spotter and a trainable rescoring head to reduce performance degradation due to the domain gap. LST-Matcher assigns IDs to text instances based on the queries in long-short term frames. The yellow star sign `{\color{yellow}{$\bigstar$}}' indicates the final output. \textbf{B. Four architectural designs of LST-Matcher}. As ST-Matcher and LT-Matcher in LST-Matcher have the same architectures, only one is shown for simplicity. \textbf{`*'} indicates that ST-Matcher and LT-Matcher shared the same network parameters.}
\label{fig:2}
\end{figure*}

\subsection{Video Text Datasets}\label{works_data}
Current progress in video text spotting research can also be attributed to the publicly accessible datasets, which serve as critical benchmarks for evaluating and advancing methods in the field.
ICDAR2015-video~\cite{Karatzas2015ICDAR15} contains 49 video clips, where the locations of words in the video are labeled using oriented bounding boxes. BOVText~\cite{wu2021bilingual} introduces a large-scale, bilingual (English and Chinese), open-world video text dataset, providing 2000+ videos with various scenarios. Recently, DSText~\cite{Wu2023DSText} establishes a dense and small text video dataset, including 100 videos. However, these datasets predominantly comprise straight text with almost no curved text, and their annotation boxes are restricted to quadrilateral shapes. To advance research in video curved text spotting, we introduce a novel benchmark comprising 60 videos, with over 30\% featuring curved text instances, designed for both training and test.

\section{Methodology}\label{med}

\subsection{Overview}\label{med_over}
The architecture of GoMatching++ is presented in Fig.~\ref{fig:2}(A). It consists of a frozen image text spotter, a rescoring head, and a Long-Short-Term Matching module (LST-Matcher).
We adopt an outstanding off-the-shelf image text spotter (\textit{i.e.}, DeepSolo) and freeze its parameters, with the aim of introducing strong text spotting capability into VTS while significantly reducing the training resources and data.
In DeepSolo, $p$ sequences of queries are used for final predictions, with each storing comprehensive semantics and spatial information for a text instance.
To alleviate spotting performance degradation caused by the image-video domain gap, we devise a rescoring mechanism, which determines the final confidence scores for text instances by leveraging both the scores from the image text spotter and a new trainable rescoring head. 
Finally, we design the LST-Matcher to generate instance trajectories by leveraging long-short-term information.

\subsection{Rescoring Mechanism}\label{med_res}
Due to the image-video domain gap, employing a frozen image text spotter for direct prediction may result in relatively low recall caused by low text confidence, further leading to a reduction in end-to-end spotting performance.
To ease this issue, we devise a rescoring mechanism via a lightweight rescoring head and a simple score fusion operation. Specifically, the trainable rescoring head is designed to recompute the score for each query from the decoder in the image text spotter. It consists of a simple linear layer and is initialized with the parameters of the image text spotter's classification head.
The score fusion operation then decides the final scores by leveraging both the scores from the image text spotter and the rescoring head.
Let $C^{t}_o=\{c^{t}_{o_1},...,c^{t}_{o_p}\}$ be a set of original scores from image text spotter in frame $t$. $C^{t}_r=\{c^{t}_{r_1},...,c^{t}_{r_p}\}$ is a set of recomputed scores obtained from the rescoring head. We obtain the maximum value for each query as the final score, denoted as $C^{t}_f=\{c^{t}_{f_1}=max(c^{t}_{o_1}, c^{t}_{r_1}),...,c^{t}_{f_p}=max(c^{t}_{o_p}, c^{t}_{r_p})\}$. \
With final scores, the queries in frames are filtered by a threshold before being sent to the LST-Matcher for association. 

\subsection{Long-Short-Term Matching Module}\label{med_lst}

\begin{figure}[t]
\includegraphics[width=0.5\textwidth]{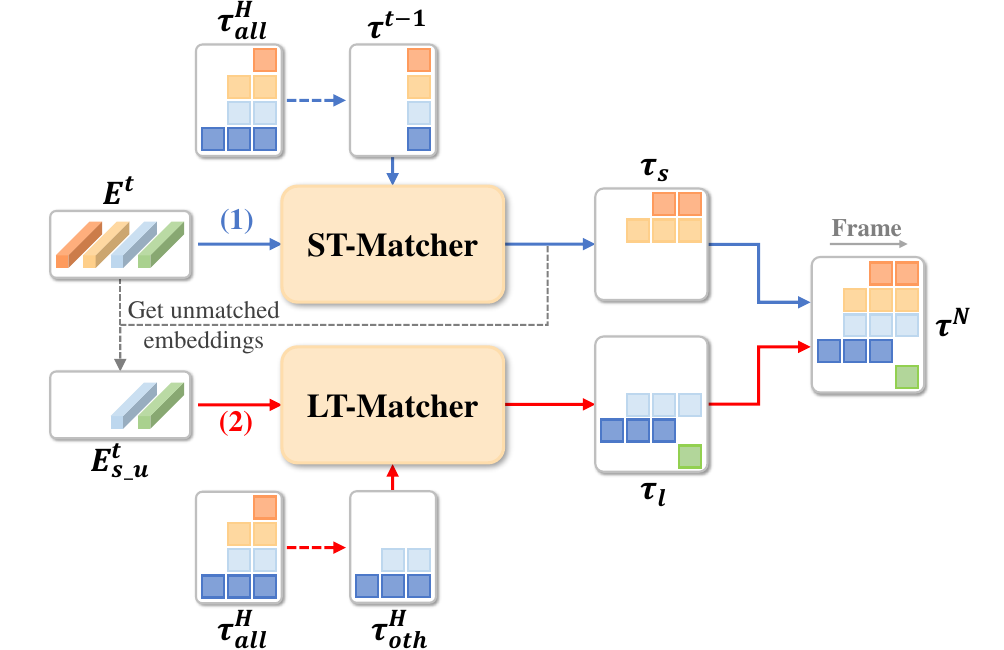} 
\caption{The inference pipeline of LST-Matcher, which is a two-stage association process: (1) ST-Matcher associates the instances with trajectories in previous frames as denoted by \textcolor[rgb]{0.3,0.5,1.0}{blue} lines. (2) LT-Matcher associates the remaining unmatched instances by utilizing other trajectories in history frames as denoted by {\color{red}{red}} lines.}
\label{fig:3}
\end{figure}

LST-Matcher comprises two sub-modules: the Short-Term Matching module (ST-Matcher) and the Long-Term Matching module (LT-Matcher), which share the same structure.
ST-Matcher is steered to match simple instances between adjacent frames into trajectories, while LT-Matcher is responsible for using long term information to address the unmatched instances due to missed detections or strong appearance changes.
We use an FFN to map the filtered text instance queries into embeddings, getting rid of using RoI features as in most existing MOT methods. The embeddings from the current frame and historical frames are processed by the LST-Matcher prior to computing the similarity score matrix. The current embeddings are then linked to the existing trajectories composed of historical embeddings or generate new trajectories according to the association score matrix.

To be specific, supposing a given clip including $T$ frames and $N_t$ text instances in frame $t$ after threshold filtering. 
$Q^t = \{q^{t}_{1},...,q^{t}_{N_t}\}$ is the set of text instance queries in frame $t$. Initially, we use a two-layer FFN to map these frozen queries into embeddings $E^t = \{e^{t}_{1},...,e^{t}_{N_t}\}$. The set of embeddings in $H$ historical frames (the input batch contains $B = H + 1$ frames) is denoted as $E^H = E^{t-H} \cup...\cup E^{t-1}$, while the corresponding trajectories are represented as $\tau^{H} = \tau^{t-H} \cup...\cup \tau^{H-1}$. And $\tau_k = \{\tau^1_k,...,\tau^{T}_k\}$ defines a spatiotemporal tube of ground truth instance locations $\tau^t_k \in \mathbbm{R}^4\cup\{\emptyset\}$, where $\tau^t_k = \emptyset$ means the absence of instance $k$ in frame $t$. During training, the matched instance indices $\hat{\alpha}^t_k$ for $\tau^t_k$ are determined based on Intersection over Union (IoU) as follows:
\begin{equation}
    \hat{\alpha}^t_k = \left\{
    \begin{array}{ll}
         \emptyset, \text{if } \tau^t_k=\emptyset \text{ or max}_i (\textit{IoU}(b^{t}_{i},\tau^t_k)) < 0.5 \\
         \text{argmax}_i (\textit{IoU}(b^{t}_{i},\tau^t_k)), \quad\text{otherwise}
    \end{array},
    \right.
    \label{eq:1}
\end{equation}
where $b^{t}_{i}$ is predicted bounding box of instance $i$ in frame $t$.

In the training phase, the ST-Matcher computes a short-term matching probabilities $G(e^t_{\hat{\alpha}^t_k}, E^{t-1})$ for $i$-th instances in frame $t$, while LT-Matcher predicts a long-term trajectory-specific matching probabilities $G(e_k, E^H)$ for $i$-th instances in frame $t$, where $e_k \in \{e^{t-H}_{\hat{\alpha}^{t-H}_k},...,e^{t}_{\hat{\alpha}^t_k}\}$. The matching probability function $G(i, j)$ is defined by the following generalized formula:
\begin{equation}
G(i, j) = \frac{\exp(S_{ij})}{\sum^N_{n=1}\exp(S_{in})}
\end{equation}

\noindent where $S_{ij}$ represents the similarity score derived from cosine similarity. Then, a cross-entropy loss function is used to optimize the predicted distributions following GTR~\cite{zhou2022global}. 

During inference, we engage a memory bank to store the instance trajectories from $H$ history frames for long term association.
All detected text instances in each frame are further processed by non-maximum-suppression (NMS) before being fed into LST-Matcher for association.
Unlike in the training phase, where ST-Matcher and LT-Matcheder are independent of each other, LST-Matcher comprises a two-stage association procedure as described in Fig.~\ref{fig:3}. Concretely, ST-Matcher first matches the embedding $E^t$ in the current frame $t$ with the trajectories $\tau^{t-1}$ in the previous frame $t-1$.
Then, LT-Matcher employs other trajectories $\tau^H_{oth}$ in the memory bank to associate the unmatched ones $E^t_{s\_u}$ with low association score in ST-Matcher caused by missed detections or strong appearance changes. 
If the association score with any trajectory calculated in ST-Matcher or LT-Matcher is higher than a threshold $\theta$, the instance is linked to the trajectory with the highest score. 
Otherwise, this instance is used to initiate a new trajectory.
Finally, we combine the trajectories $\tau_s$ and $\tau_l$ predicted by ST-Matcher and LT-Matcher to obtain new trajectories $\tau^N$ for tracking in the next frame.

\subsection{Parameter- and Data-Efficient Design}\label{med_para}
To further explore a more concise and elegant structure for the LST-Matcher, we explore four different architectures as shown in Fig.~\ref{fig:2}(B).

\textbf{(a) Transformer-based}: The ST-Matcher and LT-Matcher both incorporate a single-layer Transformer encoder and a single-layer Transformer decoder, sharing the FFN with two linear layers (denoted by `*'). This is the exact design of the GoMatching model as presented in the conference version~\cite{he2024gomatching}. The instance queries $Q^t$ from current frame $t$ are first projected into a new feature space by the FFN and obtain the embeddings $E^t$. In the encoder, the historical embeddings $E^H$ are enhanced by self-attention. The decoder takes embeddings in the current frame as query and enhanced historical embeddings as key for cross-attention, and computes the association score matrix $G$.

\textbf{(b) Similarity-only}: The instance queries $Q^t$ and $Q^H$ are used to directly compute the association score matrix $G$. However, the absence of the FFN and attention mechanism to map queries into a more suitable feature space for similarity calculation leads to a significant degradation in performance (Sec.~\ref{exp_open_abl}).

\textbf{(c) FFN-based}: The instance queries $Q^t$ are first projected by the FFN to derive the embeddings $E^t$, which are then used to compute the association score matrix $G$ with the historical embeddings $E^H$. Similarly, The absence of an attention mechanism to focus on key instance features leads to a performance decline (Sec.~\ref{exp_open_abl}).

\textbf{(d) Cross-attn-based}: The ST-Matcher and LT-Matcher are designed with a more elegant structure to obtain $G$, incorporating shared FFN and cross-attention. Notably, the number of parameters of this structure is approximately \textbf{$1/3$} that of the Transformer-based architecture (a), while maintaining comparable or even superior performance (Sec.~\ref{exp_open_abl}). 

By freezing the image text spotter and adopting the lightweight design for the proposed modules, GoMatching++ significantly reduces the number of trainable parameters, establishing a more parameter-efficient baseline. 
Additionally, the decoupled framework eliminates the need for joint optimization of detection, recognition, and tracking, enabling the matcher to fully leverage the learned features from the image text spotter. This design reduces computational overhead, stabilizes training, and enhances the use of limited data while achieving high tracking performance. Further experimental details are provided in Sec. \ref{exp_open_abl} and Sec. \ref{exp_open_cmp}.

\subsection{Optimization}\label{med_optim}
\textbf{Rescoring Loss.} To train the rescoring head, we following DETR~\cite{carion2020end} and use Hungarian algorithm~\cite{kuhn1955hungarian} to find a bipartite matching $\hat{\sigma}$ between the predictions $\hat{Y}$ and the ground truths $Y$ with minimum matching cost $\mathcal{C}$:
\begin{equation}
    \hat{\sigma} =  \arg\min\limits_{\sigma}\sum^{N}_{i}\mathcal{C}(Y_i, \hat{Y}_{\sigma(i)}),
    \label{eq:3}
\end{equation}
where \textit{N} is the number of ground truth instances per frame. The cost $\mathcal{C}$ can be defined as:
\begin{equation}
    \mathcal{C}(Y_i, \hat{Y}_{\sigma(i)}) = \lambda_c\mathcal{L}_{cls}(\hat{p}_{\sigma(i)}(c_i)) + \lambda_b\sum^N_1\Vert b_i - \hat{b}_{\sigma(i)}\Vert,
    \label{eq:4}
\end{equation}
where $\lambda_c$ and $\lambda_b$ serve as the hyper-parameters to balance different tasks. $\hat{p}_{\sigma(i)}(c_i)$ and $\hat{b}_{\sigma(i)}$ are the probability for ground truth class $c_i$ and the predicition of bounding box respectively, and $b_i$ represents the ground truth bounding box. $\mathcal{L}_{cls}$ is the focal loss~\cite{lin2017focal}. Specifically, the focal loss for training the rescoring head can be formulated as:
\begin{equation}
\begin{split}
    & \mathcal{L}_{res} = \sum^N_1 [-\mathds{1}_{\{c_i\neq\varnothing\}}\alpha(1 - \hat{p}_{\hat{\sigma}(i)}(c_i))^\gamma\log(\hat{p}_{\hat{\sigma}(i)}(c_i)) \\ 
    &- \mathds{1}_{\{c_i=\varnothing\}}(1-\alpha)(\hat{p}_{\hat{\sigma}(i)}(c_i))^\gamma\log(1-\hat{p}_{\hat{\sigma}(i)}(c_i))],
    \label{eq:5}
\end{split}
\end{equation}
where $\alpha$ and $\gamma$ are the hyper-parameters.

\textbf{Long-Short Association Loss.} In ST-Matcher, only trajectories within adjacent frames of the input batch are considered, whereas LT-Matcher extends its scope to trajectories across all frames within the batch, using long-term temporal information. For each trajectory, we optimize the log-likelihood of its assignments $\hat{\alpha}_k$ by cross-entropy, following GTR~\cite{zhou2022global}:
\begin{equation}
    \mathcal{L}_{s\_ass}=-\sum^{B}_{t=2}\log G(e^t_{\hat{\alpha}^t_k}, E^{t-1}),
    \label{eq:6}
\end{equation}

\begin{equation}
    \mathcal{L}_{l\_ass}=-\sum^{B}_{t=1}\log G(e_k, E^H),
    \label{eq:7}
\end{equation}

Finally, we can get the long-short association loss as follows:
\begin{equation}
    \mathcal{L}_{asso} = \sum_{\tau_k}(\mathcal{L}_{s\_ass} + \mathcal{L}_{l\_ass}).
    \label{eq:8}
\end{equation}

\textbf{Overall Loss.} Combined with the rescore loss $\mathcal{L}_{res}$ in Eq. (\ref{eq:5}) and the long-short association loss $\mathcal{L}_{asso}$ in Eq. (\ref{eq:8}), the final training loss can be defined as:
\begin{equation}
    \mathcal{L} = \lambda_{res}\mathcal{L}_{res} + \lambda_{asso}\mathcal{L}_{asso},
    \label{eq:9}
\end{equation}
where the hyper-parameters $\lambda_{res}$ and $\lambda_{asso}$ are the weights of $\mathcal{L}_{res}$ and $\mathcal{L}_{asso}$, respectively.

\begin{figure*}[t]
\centering
  \includegraphics[width=0.9\textwidth]{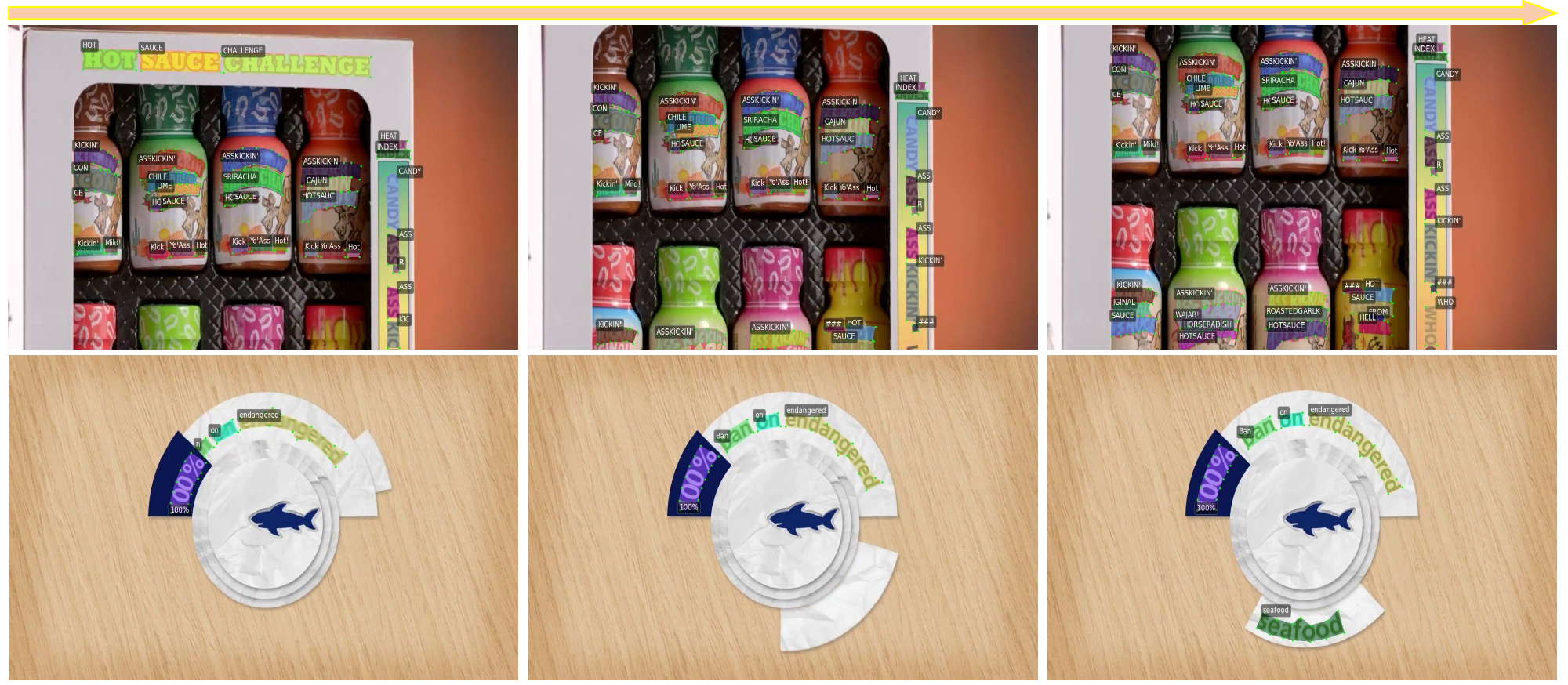}
  \caption{\textbf{Visual examples from ArTVideo training set}. The straight and curved text are labeled with quadrilaterals and polygons, respectively. The same background color in different frames denotes the same instance.}
 \label{fig:4}
\end{figure*}

\begin{table*}[t]
    \centering
    \normalsize
    \setlength{\tabcolsep}{10pt}
    \begin{tabular}{c|c|c|c|c|c|c}
\hline
 Dataset & Video & Frames & Instances & Text & Box & Mask  \\
 \hline
ICDAR15-video\cite{Karatzas2015ICDAR15} & 49 & 27k& 144k & straight & quadrilateral& $\times$ \\
BOVText\cite{wu2021bilingual} & 2,021 & 1.7m& 8.8m & straight& quadrilateral&$\times$ \\
DSText\cite{Wu2023DSText} & 100 & 56k& 671k& straight& quadrilateral&$\times$ \\
\hline
\multirow{3}{*}{ArTVideo}
 & & & 170k& & &  \\
 & 60 & 13k& (58k curved text, & arbitrary & polygon& $\checkmark$ \\
 & & & 112k straight text) & & &  \\
\hline
\end{tabular}
    \caption{\textbf{Comparison with existing video text datasets.} `Text' and `Box' denote the types of text instances and annotation formats, respectively, while `Mask' indicates whether each text instance includes mask annotations.}
    \label{table:1}
\end{table*}

\section{The ArTVideo Benchmark}\label{art}
The scarcity of curved text instances in existing video text spotting datasets renders it significantly challenging for both the development and evaluation of video text spotting models in scenarios involving curved text.
To fill this gap, we establish a novel benchmark named ArTVideo, encompassing a total of 60 videos with over 30\% curved text. In this section, we provide a comprehensive overview of the benchmark and its characteristics.

\subsection{Data Source and Annotation}\label{art_data}
\textbf{Data Source.}
The ArTVideo dataset consists of 60 curved text videos with a total of 12,711 frames, sourced from existing video datasets and publicly available YouTube videos. Specifically, 2 videos are from ICDAR15-video, 3 from BOVText, and 55 from YouTube. The dataset includes 169,802 text instances, with 111,784 straight text and 58,018 curved text, making curved text approximately 34\% of the total. The videos span diverse scenarios, including driving, short-form content, outdoor scenes, product showcases, advertisements, and educational materials.

\begin{figure*}[t]
\centering
  \includegraphics[width=0.8\textwidth]{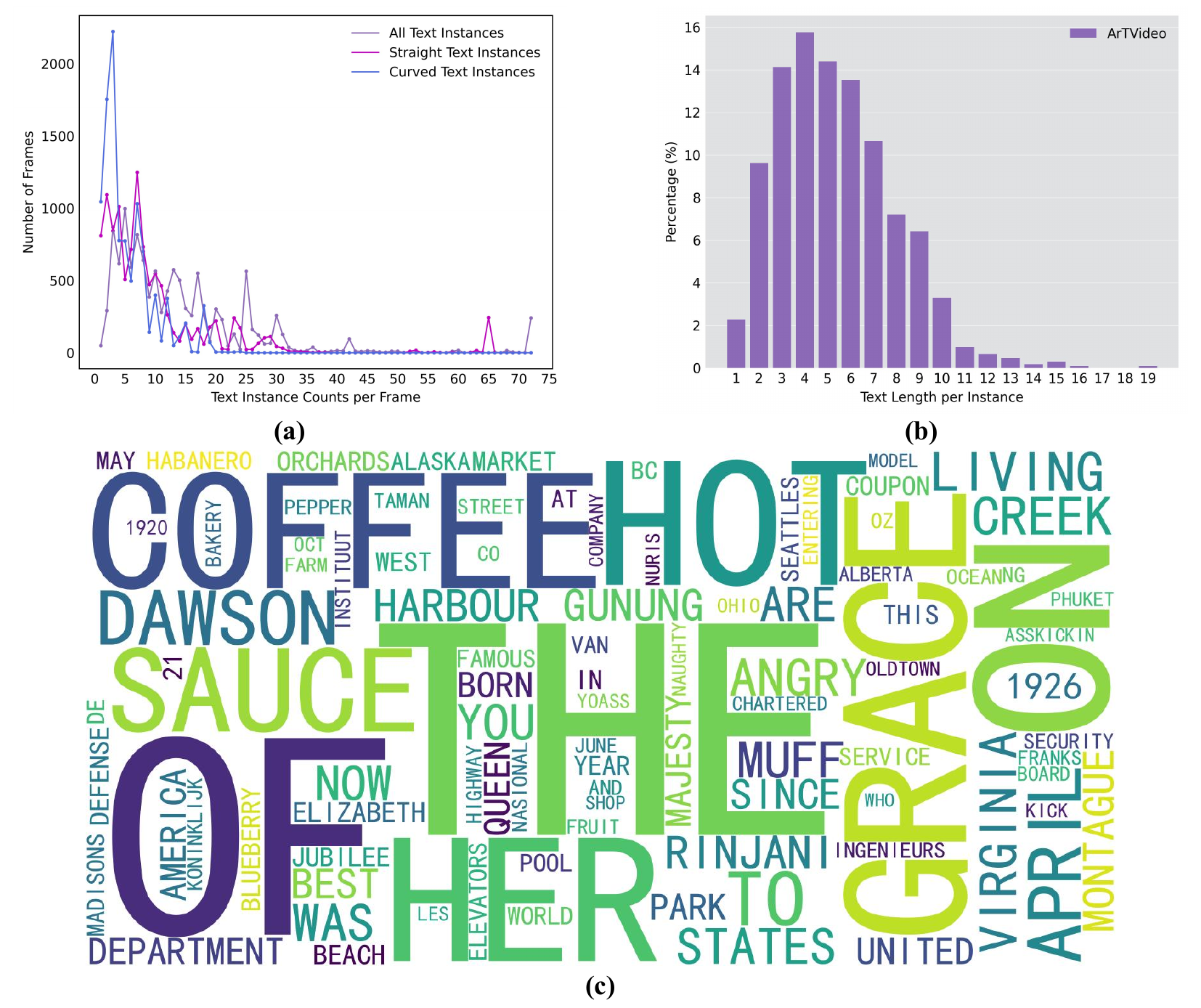}
  \caption{\textbf{Statistics of ArTVideo.} (a) and (b) show the distribution of various types of text instance numbers in each frame and the distribution of the text length of each instance, respectively. (c) presents the word cloud of text annotations in ArTVideo.}
 \label{fig:5}
\end{figure*}

\noindent\quad\textbf{Data Annotation.}
As shown in Fig.~\ref{fig:4}, we provide high-quality word-level annotations for both straight and curved text in two different annotated styles. Straight text is labeled with quadrilaterals, while for curved text, we follow the CTW1500~\cite{liu2019curved} and adopt a polygon with 14 points to annotate the text contour. To record the details of each text instance, we annotate the following attributes: 1) \textit{ID} refers to the unique identifier assigned to each text instance across consecutive frames; 2) \textit{Text Category} is classified into two types: `alphanumeric', which includes Latin text, and `other', which encompasses both blurry and non-Latin text; 3) \textit{Transcription} means the text string for each text region; 4) \textit{Bbox} and \textit{Poly} represent the bounding boxes of text instances, in horizontal and polygonal formats, respectively; 5) \textit{Segmentation} denotes the mask annotation for each text region; 6) \textit{Box Type} indicates the type of box annotation for each text instance, which is either `quadrilateral' (for straight text) or `polygon' (for curved text).

The labor-intensive labeling process, which included annotation and cross-verification to ensure quality, involved a team of 8 members working for half a month, amounting to approximately 960 man-hours to annotate 60 videos.

\subsection{Statistical Analysis} \label{art_stat}
\textbf{Dataset Characteristics.}
The details of ArTVideo, along with a comparison to other existing video text datasets, are presented in Tab.~\ref{table:1}. The table clearly highlights three key characteristics of ArTVideo: 1) It contains over 34\% curved text, whereas other datasets primarily consist of straight text; 2) The annotation boxes are in polygonal form, enhancing annotation precision; 3) ArTVideo includes additional mask annotations, allowing models trained on this dataset to produce more precise segmentation results for text regions. These distinctive characteristics make ArTVideo a valuable resource for advancing video text spotting research, particularly in scenarios involving curved text and the need for precise segmentation results.

\noindent\quad\textbf{Statistical Analysis.}
In Fig.~\ref{fig:5} (a) and (b), we provide statistical information about ArTVideo, including the distribution of various types of text instances per frame—namely all text, straight text, and curved text—as well as the distribution of text lengths for each text instance. From panel (a), it is evident that most frames in ArTVideo contain approximately 1 to 15 curved text instances and 1 to 20 straight text instances, resulting in the total text instance counts per frame predominantly falling within the range of 2 to 30. In rare cases, some individual frames contain over 70 text instances, highlighting the presence of scenes with dense text regions. Moreover, as illustrated in panel (b), most text instances have lengths concentrated within the range of 2 to 10 characters, while a small proportion of instances extending beyond this range, with the longest reaching up to 19 characters. This suggests that while short text instances are prevalent, the dataset also includes longer instances, which represent more complex or detailed textual information in certain scenes.

\noindent\quad\textbf{Word Cloud.}
We present a word cloud visualization of the textual content in Fig.~\ref{fig:5} (c). The most frequently occurring word in this dataset is `THE' followed by `OF', `ON', `HOT', `HER', and other words with lengths of no more than 10 characters. These observations are consistent with the statistical results of the text length distribution shown in panel (b).

\subsection{Supported Tasks and Metrics}\label{art_sup}
Since the text instances in our ArTVideo are arbitrary-shape, it differs from previous datasets that primarily focus on video text tracking and video text spotting. Our benchmark additionally supports two distinctive tasks: video curved text tracking and video curved text spotting, both of which are specifically designed to evaluate model performance on curved text instances.

Following ICDAR15-video~\cite{Karatzas2015ICDAR15}, BOVText~\cite{wu2021bilingual} and DSText~\cite{Wu2023DSText}, the tracking and spotting performance of the models are assessed with three metrics: \textit{MOTA} (Multiple Object Tracking Accuracy)~\cite{bernardin2008evaluating}, \textit{MOTP} (Multiple Object Tracking Precision), and \textit{IDF1}~\cite{ristani2016performance}. \textit{MOTA} indicates the overall performance of detection and tracking but prefers to reflect detection performance. \textit{MOTP} measures the average spatial distance between the predicted locations of tracked objects and the ground truth positions. \textit{IDF1} provides a measurement of association accuracy. 

\section{Experiments on Open-source Datasets}\label{exp_open}

\subsection{Datasets and Evaluation Metrics}\label{exp_open_data}
\textbf{ICDAR13-video}~\cite{karatzas2013icdar} contains 13 videos for training and 15 videos for testing from indoors and outdoors scenarios with quadrilateral annotations at word-level.

\noindent\quad\textbf{ICDAR15-video}~\cite{Karatzas2015ICDAR15} is an extended version of ICDAR2013-video, serving as a word-level video text benchmark with quadrilateral annotations, comprising 25 training and 24 test videos, focusing on wild scenarios.

\noindent\quad\textbf{BOVText}~\cite{wu2021bilingual} is a bilingual video text spotting benchmark (English and Chinese) with textline-level quadrilateral annotations, covering different scenarios in 2,021 videos, such as `Cartoon', `Vlog', `Movie', and more.

\noindent\quad\textbf{DSText}~\cite{Wu2023DSText} provides 50 training videos and 50 test videos. Compared with the previous datasets, DSText mainly includes the following three new challenges: dense video texts, high-proportioned small texts, and various new scenarios. Similar to ICDAR15-video, DSText adopts word-level annotations, which are labeled with quadrilaterals.

\noindent\quad\textbf{Evaluation Metrics.} To evaluate the performance of methods, we adopt three evaluation metrics commonly used in previous benchmarks, \textit{i.e.}, MOTA, MOTP, and IDF1.

\begin{table*}[t]
\begin{minipage}[t]{\columnwidth}
\captionsetup{type=table}
\small
\centering
\setlength{\tabcolsep}{2pt}
\caption{\textbf{Impact of difference components in the proposed GoMatching++.}
}

\begin{tabular}{c|cc|ccc}
\hline
Index &ReScoring &Matcher &MOTA ($\uparrow$) &MOTP ($\uparrow$) &IDF1 ($\uparrow$) \\
\hline
1 & &LST-Matcher &70.68 &\textbf{78.73} &78.42 \\
2 &$\checkmark$ &ST-Matcher &71.17 &78.41 &75.53 \\
3 &$\checkmark$ &LT-Matcher &70.07 &78.52 &78.48 \\
4 &$\checkmark$ &LST-Matcher &\textbf{72.20} &78.52 &\textbf{80.11} \\
\hline
\end{tabular}
\label{table:2}
\end{minipage}
\hfill
\begin{minipage}[t]{\columnwidth}
\captionsetup{type=table}
    \setlength{\tabcolsep}{4pt}
    \small 
    \centering
    \caption{\textbf{Video text detection performance on ICDAR2013-video.} The best results are highlighted in \textbf{bold}.}
    \begin{tabular}{cccc}
    \hline
    Method &Prec. ($\uparrow$) &Rec. ($\uparrow$) &F-Score ($\uparrow$) \\
    \hline
    Free~\cite{cheng2020free} &79.7 &68.4 &73.6 \\
    TransDETR~\cite{wu2024end} &80.6  &70.2  &75.0 \\
    GoMatching++ w/o rescoring &\textbf{92.4}  &65.7  &76.8 \\
    GoMatching++  &89.5  &\textbf{74.8}  &\textbf{81.5} \\
    \hline
    \end{tabular}
    \label{table:3}
\end{minipage}
\end{table*}

\begin{table*}[t]
    \setlength{\tabcolsep}{3pt}
    \small 
    \centering
    \caption{\textbf{Video text spotting results for different LST-Matcher architectural designs}. `T-Para' denotes the number of trainable parameters.
    }
    \begin{tabular}{c|ccc|ccc|ccc|c}
    \hline
    \multirow{2}{*}{Architectures} & \multicolumn{3}{c|}{ICDAR15-Video}  &\multicolumn{3}{c|}{BOVText} & \multicolumn{3}{c|}{DSText}  & \multirow{2}{*}{T-Para (M)} \\ \cline{2-10}
    &MOTA ($\uparrow$) &MOTP ($\uparrow$) &IDF1 ($\uparrow$) &MOTA ($\uparrow$) &MOTP ($\uparrow$) &IDF1 ($\uparrow$) &MOTA ($\uparrow$) &MOTP ($\uparrow$) &IDF1 ($\uparrow$) & \\
    \hline
    Transformer-based &\underline{72.04}  &\textbf{78.53}  &\textbf{80.11} &\textbf{52.9}  &87.2  &\underline{62.6} &\underline{22.83}  &\underline{80.43}  &\underline{46.09} & 32.79 \\
    Similarity-only &68.64  &78.44  &72.68 &52.8  &87.2  &57.4 &17.23  &\textbf{80.49}  &39.35 & -  \\
    FFN-based &70.20  &78.50  &77.83 &\textbf{52.9}  &87.2  &61.0 &21.25  &\underline{80.43}  &44.80 & 7.60 \\
    Cross-attn-based &\textbf{72.20}  &\underline{78.52}  &\textbf{80.11} &\textbf{52.9}  &87.2  &\textbf{62.8} &\textbf{23.23}  &80.42  &\textbf{46.24} & 11.80 \\
    \hline
    \end{tabular}
    \label{table:6}
\end{table*}

\subsection{Implementation Details}\label{exp_open_det}
In all experiments, we only use a single NVIDIA GeForce RTX 3090 (24G) GPU to train and evaluate GoMatching and GoMatching++. As for the image text spotter, we apply the officially released DeepSolo~\cite{ye2023deepsolo}. To adapt GoMatching and GoMatching++ for downstream video datasets, we only update the rescoring head and LST-Matcher, while keeping DeepSolo frozen. 

\textbf{Training Setting.} 
The image text spotting part is initialized with off-the-shelf DeepSolo weights and remains frozen in all experiments. We optimize other modules on video datasets.
We follow EfficientDet~\cite{tan2020efficientdet} to adopt the scale-and-crop augmentation strategy with a resolution of 1,280. The batch size $B$ is 6. All frames in a batch are from the same video.
AdamW~\cite{loshchilov2018decoupled} is used as the optimizer. We adopt the warmup cosine annealing learning rate strategy with the initial learning rate being set to $5\times10^{-5}$. 
The loss weights $\lambda_{res}$ and $\lambda_{asso}$ are set to 1.0 and 0.5, respectively. For focal loss, $\alpha$ is 0.25 and $\gamma$ is 2.0 as in~\cite{carion2020end,ye2023deepsolo}. The model is trained for 30k iterations on all downstream video datasets.

\noindent\quad\textbf{Inference Setting.} The inference settings of GoMatching and GoMatching++ are identical, ensuring a fair and reliable comparison of their performance. The association score threshold is set to 0.2. For ICDAR13-video and ICDAR15-video, the shorter size of the input image is resized to 1,000. Except explicitly stated, ablation studies are conducted using a shorter side size of 1,440 on ICDAR15-video, which aligns with the configuration used in DeepSolo. As for BOVText and DSText, the shorter sizes are set to 1,000 and 1,280, respectively. The number of history frames in the tracking memory bank is 5. Additionally, trajectories shorter than five frames are discarded to enhance the robustness of the tracking.

\subsection{Ablation Studies}\label{exp_open_abl}
\noindent\quad\textbf{Effectiveness of Rescoring Mechanism.}
As presented in row 1 and row 4 of Tab.~\ref{table:2}, the rescoring mechanism can effectively combining the knowledge of rescoring head learned from the video dataset with the prior knowledge of DeepSolo, resulting performance gains of 1.52\% and 1.69\% on MOTA and IDF1, respectively. 
Moreover, following previous works~\cite{cheng2020free, wu2024end}, we evaluate the performance of GoMatching++ for video text detection on ICDAR13-video. Since GoMatching and GoMatching++ employ the same text detector, we only report the video text detection performance of GoMatching++. As shown in Tab.~\ref{table:3}, without the rescoring mechanism, GoMatching++ depends on the original outputs from DeepSolo, resulting in a 4.5\% decrease in recall and only a 1.8\% improvement in F-Score compared to TransDETR. This performance drop is due to the domain gap between image and video data, as directly applying an image text spotter yields low-confidence results and poor recall on video data. However, with the rescoring mechanism, GoMatching++ achieves a 9.1\% improvement in recall over DeepSolo and a 6.5\% boost in F-Score over TransDETR. 
These gains highlight the effectiveness of the rescoring mechanism in mitigating the domain gap and improving the tracking candidate pool, demonstrating that GoMatching++'s strong performance is not solely due to the use of a robust image text spotter.

\begin{figure*}[t]
\centering
\resizebox{0.9\textwidth}{!}{
\begin{tabular}{cccc}
  \includegraphics[width=0.25\textwidth]{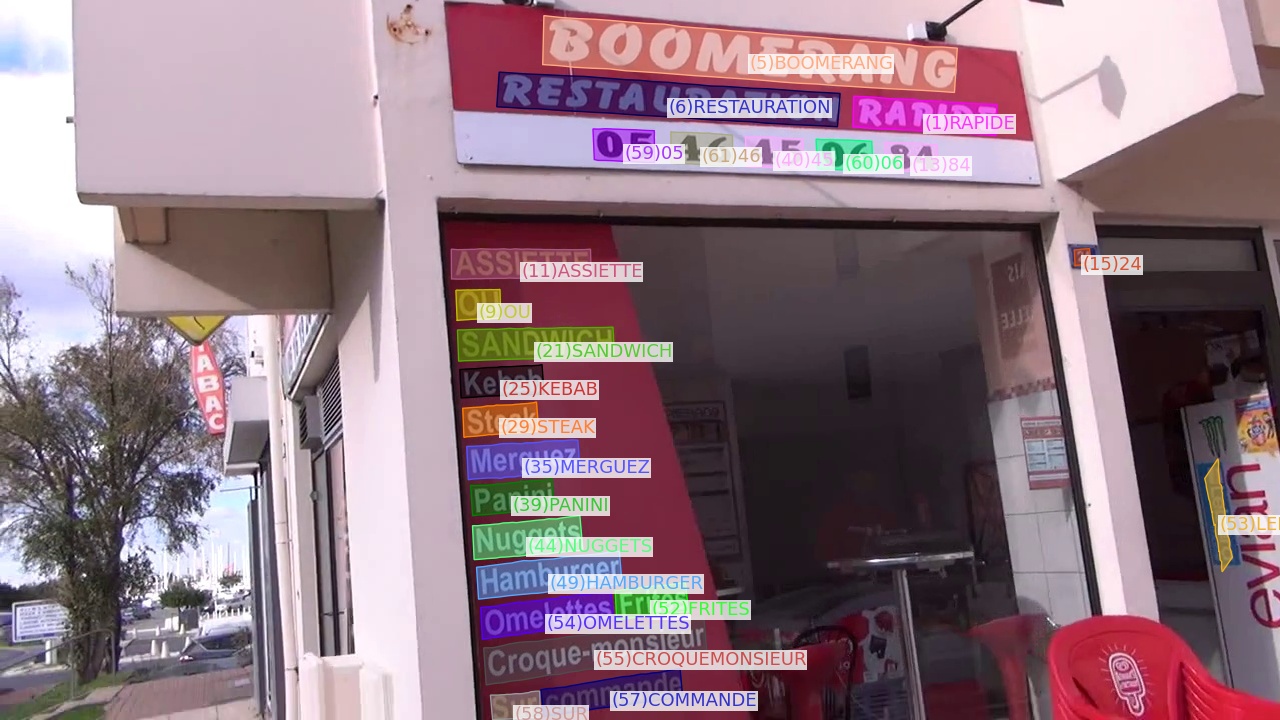} 
  &\includegraphics[width=0.25\textwidth]{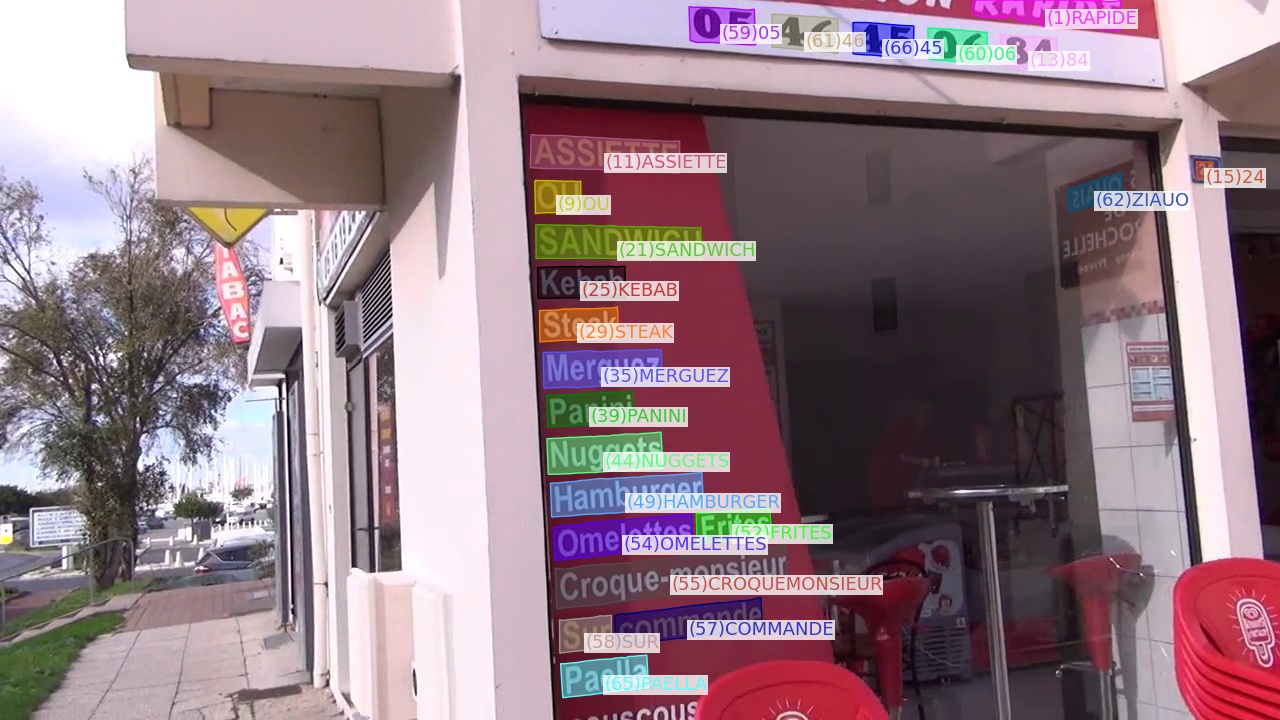}
  &\includegraphics[width=0.25\textwidth]{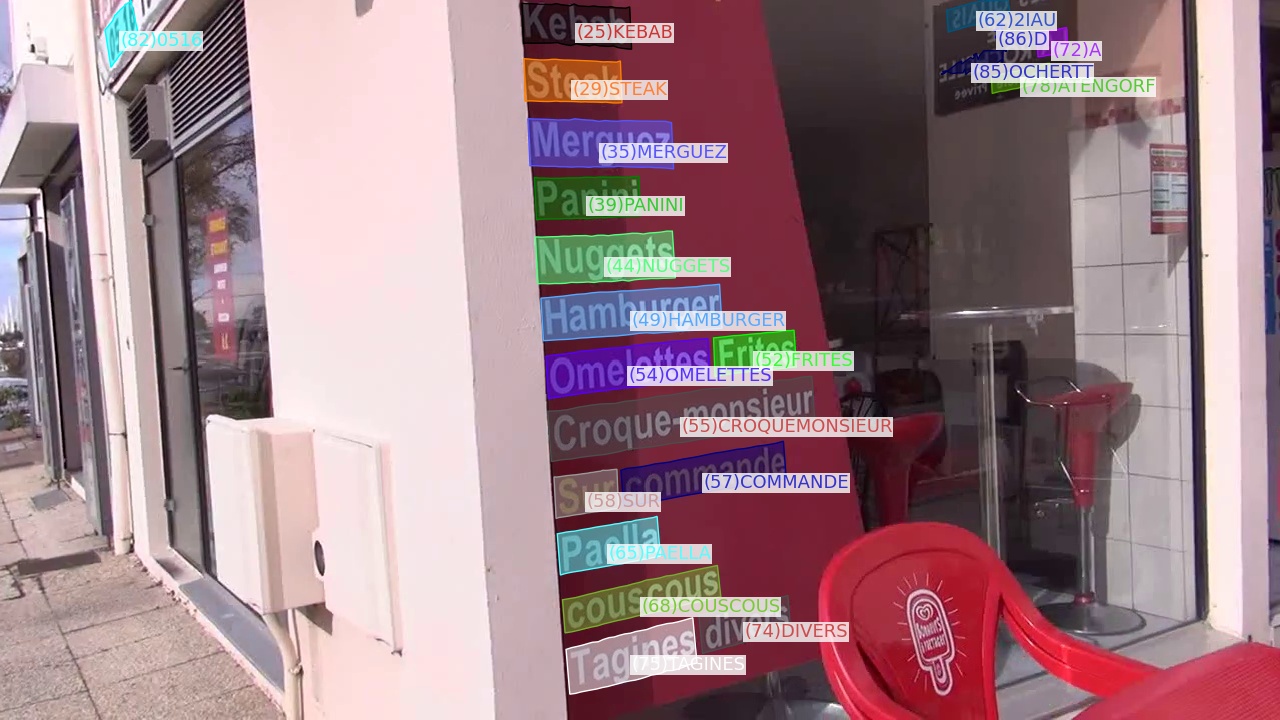}
  &\includegraphics[width=0.25\textwidth]{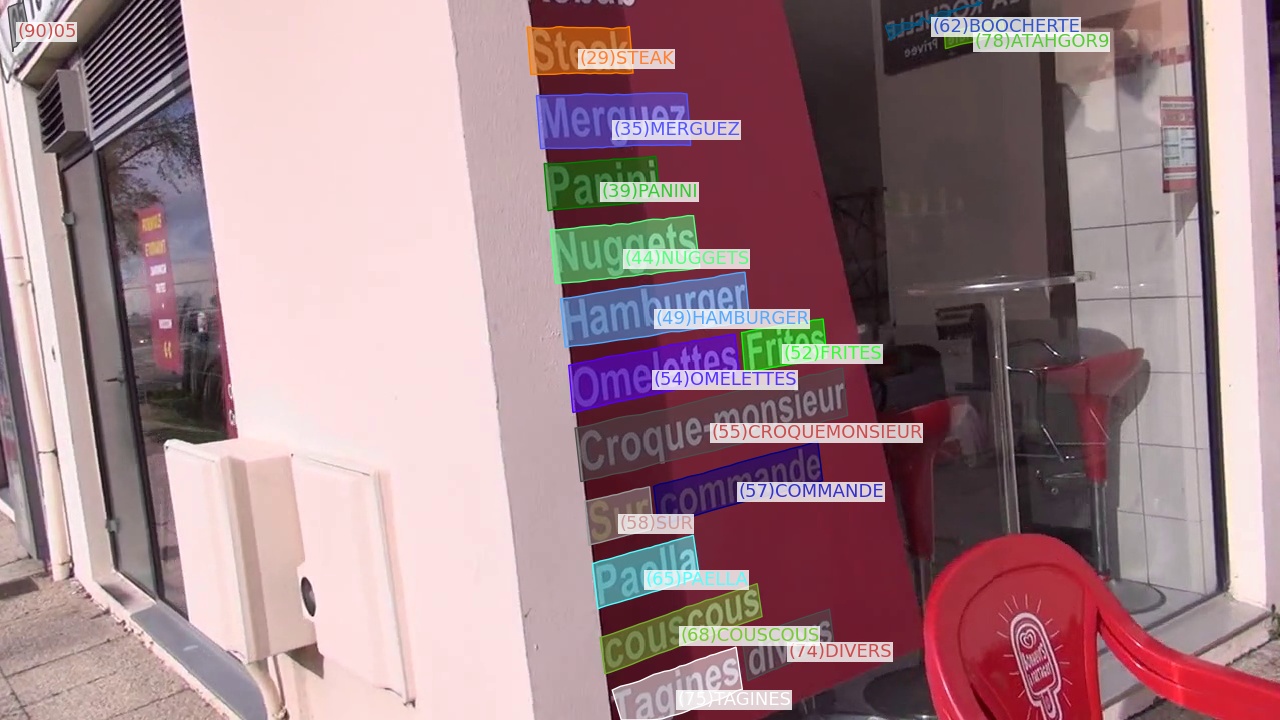}
  \\ 
  \includegraphics[width=0.25\textwidth]{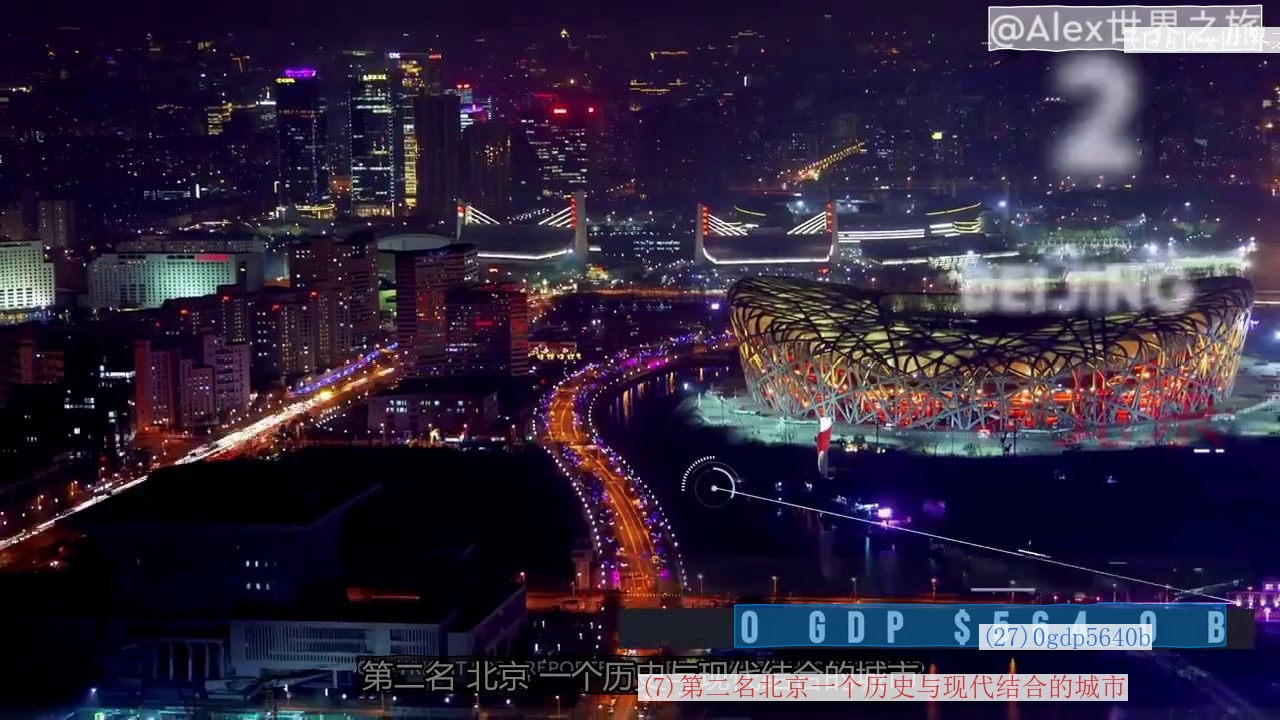} 
  &\includegraphics[width=0.25\textwidth]{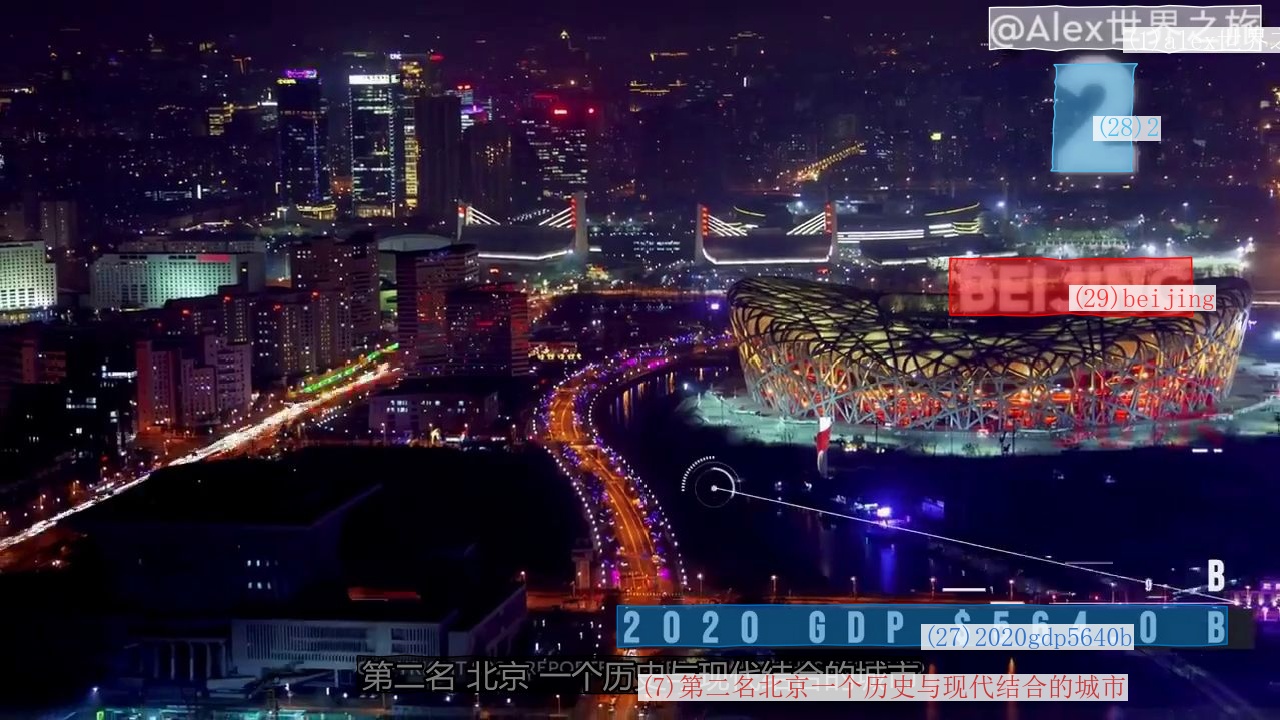}
  &\includegraphics[width=0.25\textwidth]{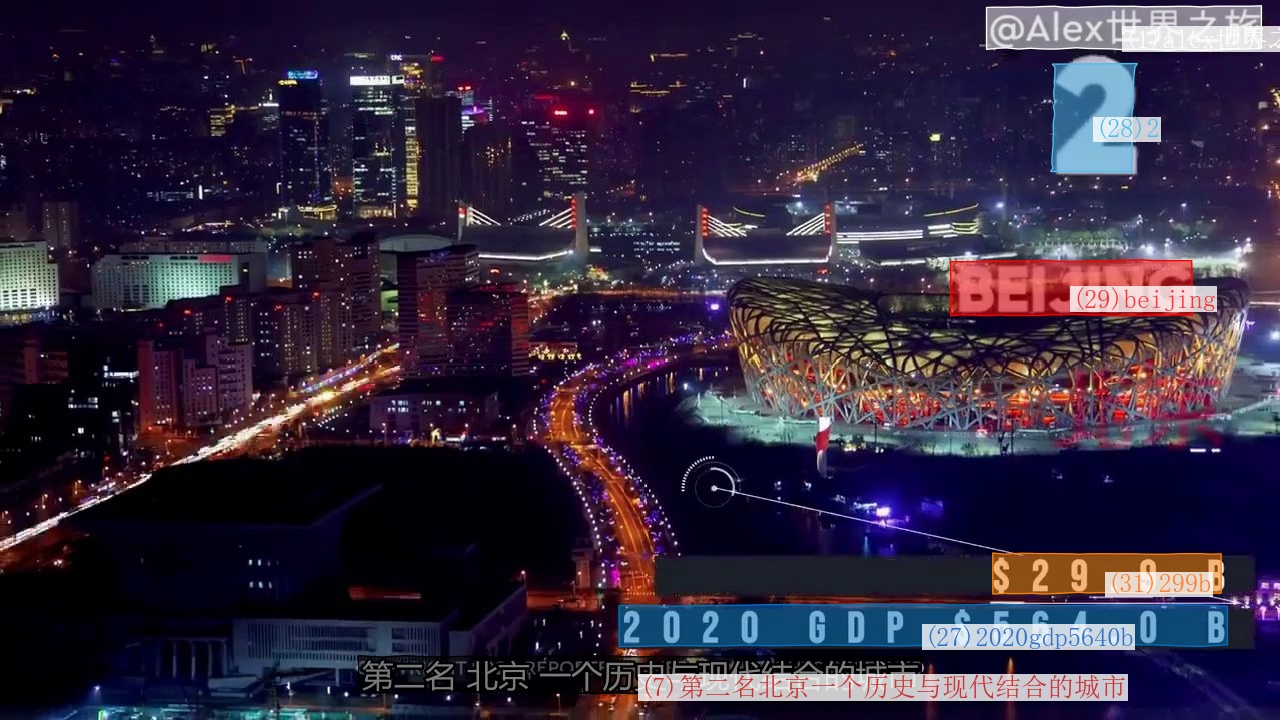}
  &\includegraphics[width=0.25\textwidth]{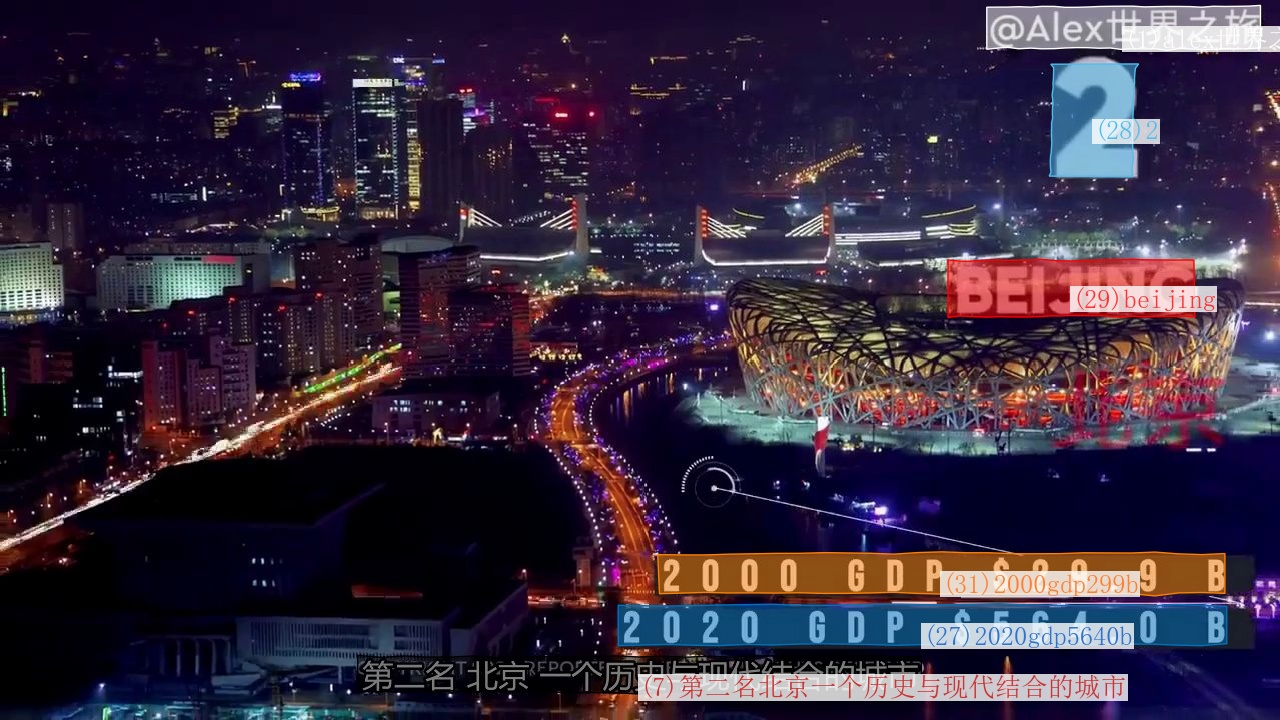}
  \\
  \includegraphics[width=0.25\textwidth]{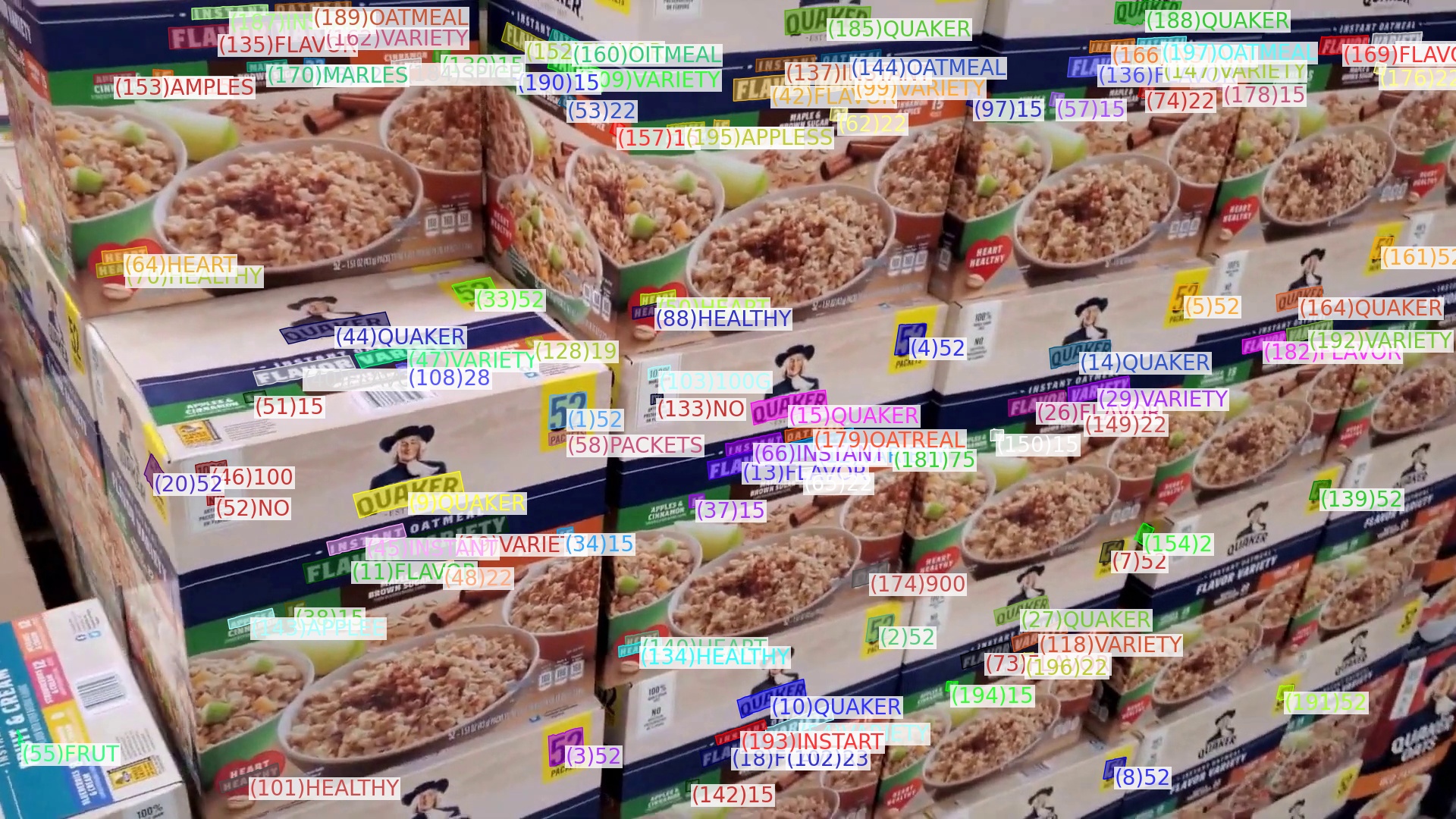}
  &\includegraphics[width=0.25\textwidth]{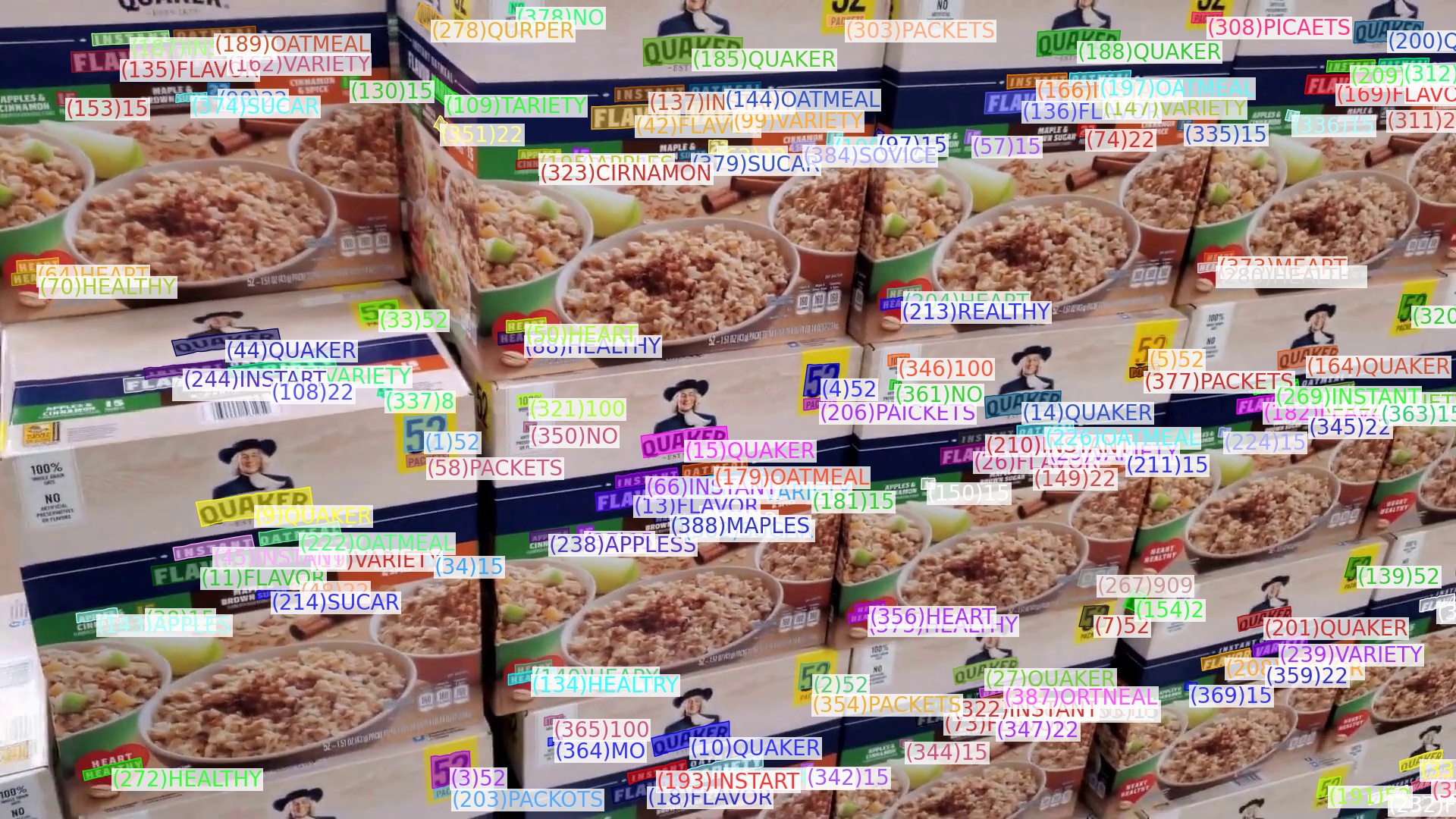}
  &\includegraphics[width=0.25\textwidth]{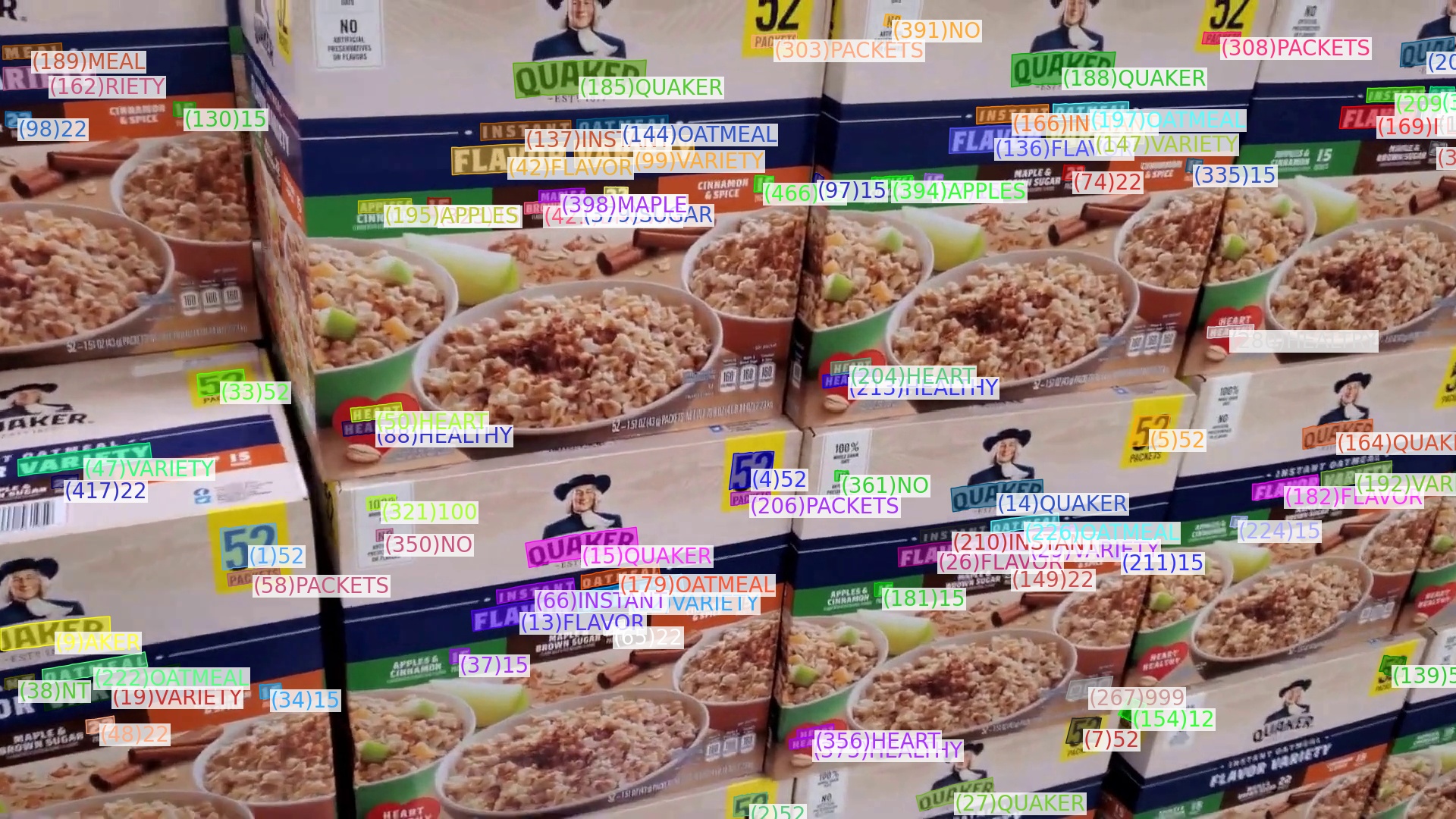}
  &\includegraphics[width=0.25\textwidth]{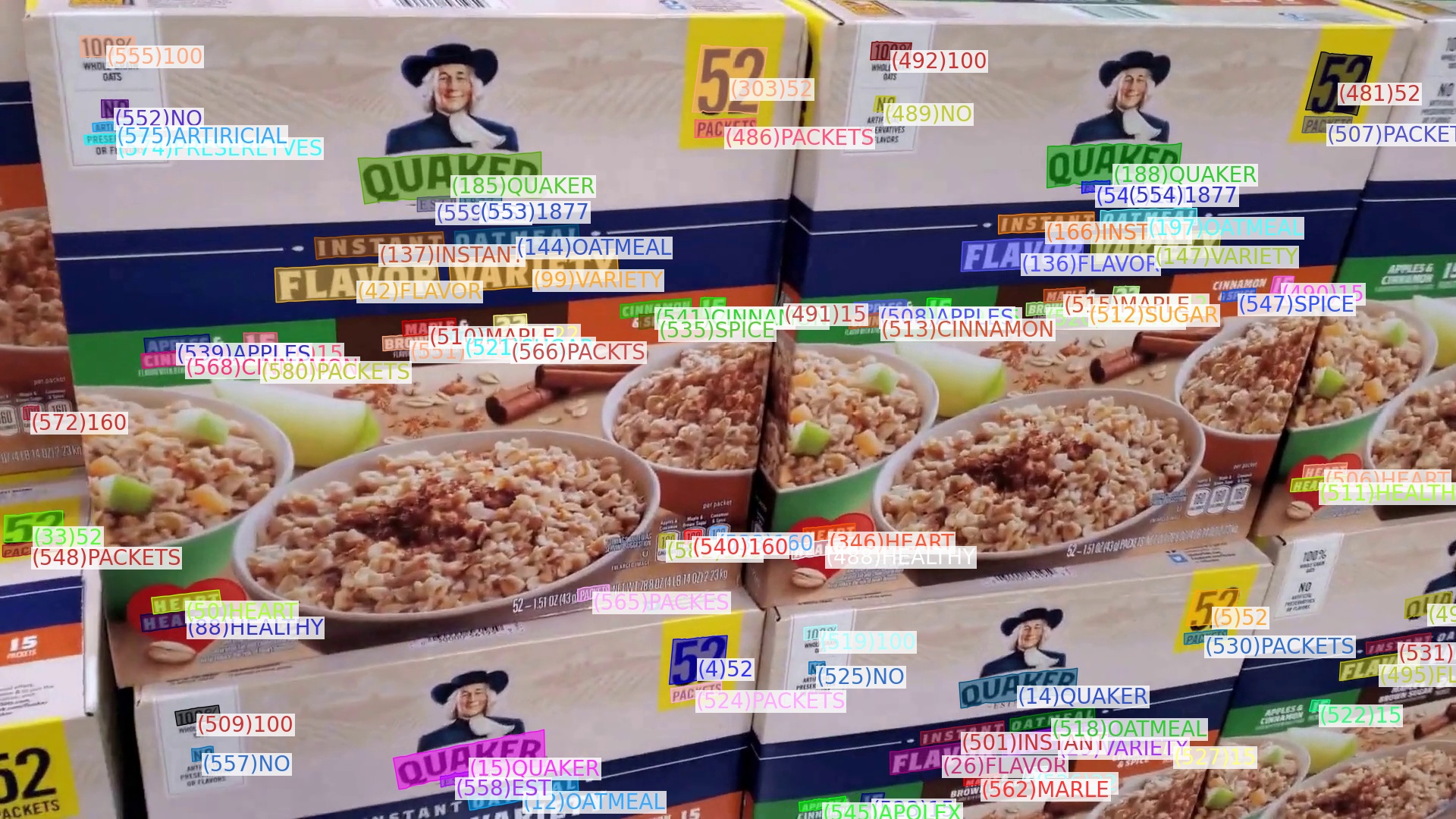}
  \\
  \end{tabular}}
\caption{
\textbf{Visual results of video text spotting achieved by GoMatching++.} From top to bottom: results on ICDAR15-video, BOVText, and DSText. Text instances associated with the same trajectories are assigned the same colors.}
 \label{fig:7}
\end{figure*}

\begin{table*}[t]
    \setlength{\tabcolsep}{5.5pt}
    \small 
    \centering
    \caption{
    \textbf{Comparison with SOTA methods on three open-source datasets.} `\dag' indicates results from the official competition website, and `*' denotes results from the officially released model. `M-ME' shows whether multi-model ensembling was used (`Y' for Yes, `N' for No). The best and second-best results are marked in \textbf{bold} and \underline{underlined}, respectively.
    }
    \begin{tabular}{c|c|ccc|ccc|c}
    \hline
    \multirow{2}{*}{Dataset} & \multirow{2}{*}{Method} & \multicolumn{3}{c|}{Video Text Tracking}  &\multicolumn{3}{c|}{Video Text Spotting}  & \multirow{2}{*}{M-ME} \\ \cline{3-8}
     & &MOTA ($\uparrow$) &MOTP ($\uparrow$) &IDF1 ($\uparrow$) &MOTA ($\uparrow$) &MOTP ($\uparrow$) &IDF1 ($\uparrow$) \\
    \hline
    \multirow{8}{*}{ICDAR15-video}
    & HIK\_OCR~\cite{cheng2020free} &43.16 &76.78 &57.92 &52.98 &74.88 &61.85 & N \\
    & CoText~\cite{wu2024bilingual} &50.3 &73.6 &67.4  &58.94  &74.53  &71.66 & N \\
    & TransDETR~\cite{wu2024end} &48.86 &74.01 &65.33 &60.96 &74.61 &72.80 & N\\
    & CoText(Kuaishou\_MMU)\dag & - & - & - &66.96  &76.55  &74.24 & - \\
    &\cellcolor{gray!20}GoMatching (ours) &\cellcolor{gray!20}\underline{62.24} &\cellcolor{gray!20}\textbf{77.94} &\cellcolor{gray!20}\underline{72.81} &\cellcolor{gray!20}\underline{72.04}  &\cellcolor{gray!20}\textbf{78.53}  &\cellcolor{gray!20}\textbf{80.11} &\cellcolor{gray!20}N \\
    &\cellcolor{gray!20}GoMatching++ (ours) &\cellcolor{gray!20}\textbf{62.41} &\cellcolor{gray!20}\textbf{77.94} &\cellcolor{gray!20}\textbf{72.87} &\cellcolor{gray!20}\textbf{72.20}  &\cellcolor{gray!20}\underline{78.52}  &\cellcolor{gray!20}\textbf{80.11} &\cellcolor{gray!20}N
    \\
    \hline
    \multirow{6}{*}{BOVText}
    & TransVTSpotter~\cite{wu2021bilingual} &68.2 &82.1 & 64.7 &-1.4  &\underline{82.0}  &40.8 & N \\
    & CoText~\cite{wu2024bilingual} &69.7 & 80.3 & \textbf{76.5}
 &10.3  &80.4  &47.1 & N \\
     &\cellcolor{gray!20}GoMatching (ours) &\cellcolor{gray!20}\textbf{70.2} &\cellcolor{gray!20}\textbf{86.7} &\cellcolor{gray!20}69.7 &\cellcolor{gray!20}\textbf{52.9}  &\cellcolor{gray!20}\textbf{87.2}  &\cellcolor{gray!20}\underline{62.6}  &\cellcolor{gray!20}N \\
     &\cellcolor{gray!20}GoMatching++ (ours) &\cellcolor{gray!20}\textbf{70.2} &\cellcolor{gray!20}\textbf{86.7} &\cellcolor{gray!20}\underline{70.0}  &\cellcolor{gray!20}\textbf{52.9}  &\cellcolor{gray!20}\textbf{87.2}  &\cellcolor{gray!20}\textbf{62.8}  &\cellcolor{gray!20}N\\
    \hline
    \multirow{8}{*}{DSText}
    & TransDETR+HRNet\dag &43.52 &78.15 &62.27 &-28.58 &80.36 &26.20 &Y \\
    & DA\dag &\underline{50.52} &78.33 &70.99 &10.51  &78.97  &\underline{53.45} &Y \\
    & TencentOCR\dag &\textbf{62.56} &79.88 &\textbf{75.87} &22.44  &\textbf{80.82}  &\textbf{56.45} &Y \\
    & TransDETR~\cite{wu2024end}* &38.98 &77.15 &55.61 &-22.63 &79.73 &26.43 &N \\
     &\cellcolor{gray!20}GoMatching (ours) &\cellcolor{gray!20}47.96 &\cellcolor{gray!20}\textbf{80.01} &\cellcolor{gray!20}58.40 &\cellcolor{gray!20}\underline{22.83}  &\cellcolor{gray!20}\underline{80.43}  &\cellcolor{gray!20}46.09 &\cellcolor{gray!20}N \\
     &\cellcolor{gray!20}GoMatching++ (ours) &\cellcolor{gray!20}48.02 &\cellcolor{gray!20}\textbf{80.01} &\cellcolor{gray!20}58.44 &\cellcolor{gray!20}\textbf{23.23}  &\cellcolor{gray!20}80.42  &\cellcolor{gray!20}46.24 &\cellcolor{gray!20}N \\
    \hline
    \end{tabular}
    \label{table:7}
\end{table*}

\noindent\quad\textbf{Effectiveness of LST-Matcher.}
In this part, we conduct three experimental settings to evaluate the effectiveness of the LST-Matcher. ST-Matcher associates text instances in the current frame with trajectories from the previous frame, while LT-Matcher associates text instances in the current frame with trajectories stored in the tracking memory bank. The LST-Matcher combines both ST-Matcher and LT-Matcher.
As shown in rows 2–4 of Tab.~\ref{table:2}, LST-Matcher outperforms ST-Matcher by 1.03\% in MOTA and 4.58\% in IDF1, and surpasses LT-Matcher by 2.13\% in MOTA and 1.63\% in IDF1.
The results demonstrate that integrating short-term and long-term information can improves tracking robustness, thereby enhancing video text spotting performance.

\noindent\quad\textbf{Different Architectures of LST-Matcher.}
We investigate optimal architectural designs for LST-Matcher, with results summarized in Tab.~\ref{table:6}. For fairly performance comparison across variants, we maintain consistent structure and parameters for all other components. Tab.~\ref{table:6} shows that removing either FFN, which maps queries to a more suitable feature space, or the attention mechanism, which focuses on key instance features, leads to a significant drop in performance. Notably, a simplified design, which reducing the attention mechanism to cross-attention only, achieves a two-thirds reduction in parameters while maintaining or even surpassing the performance of transformer-based architectures across all datasets. This simple yet effective design highlights the strong potential of GoMatching++ as a parameter-efficient baseline for video text spotting.

\begin{figure*}[t]
\centering
\includegraphics[width=0.8\textwidth]{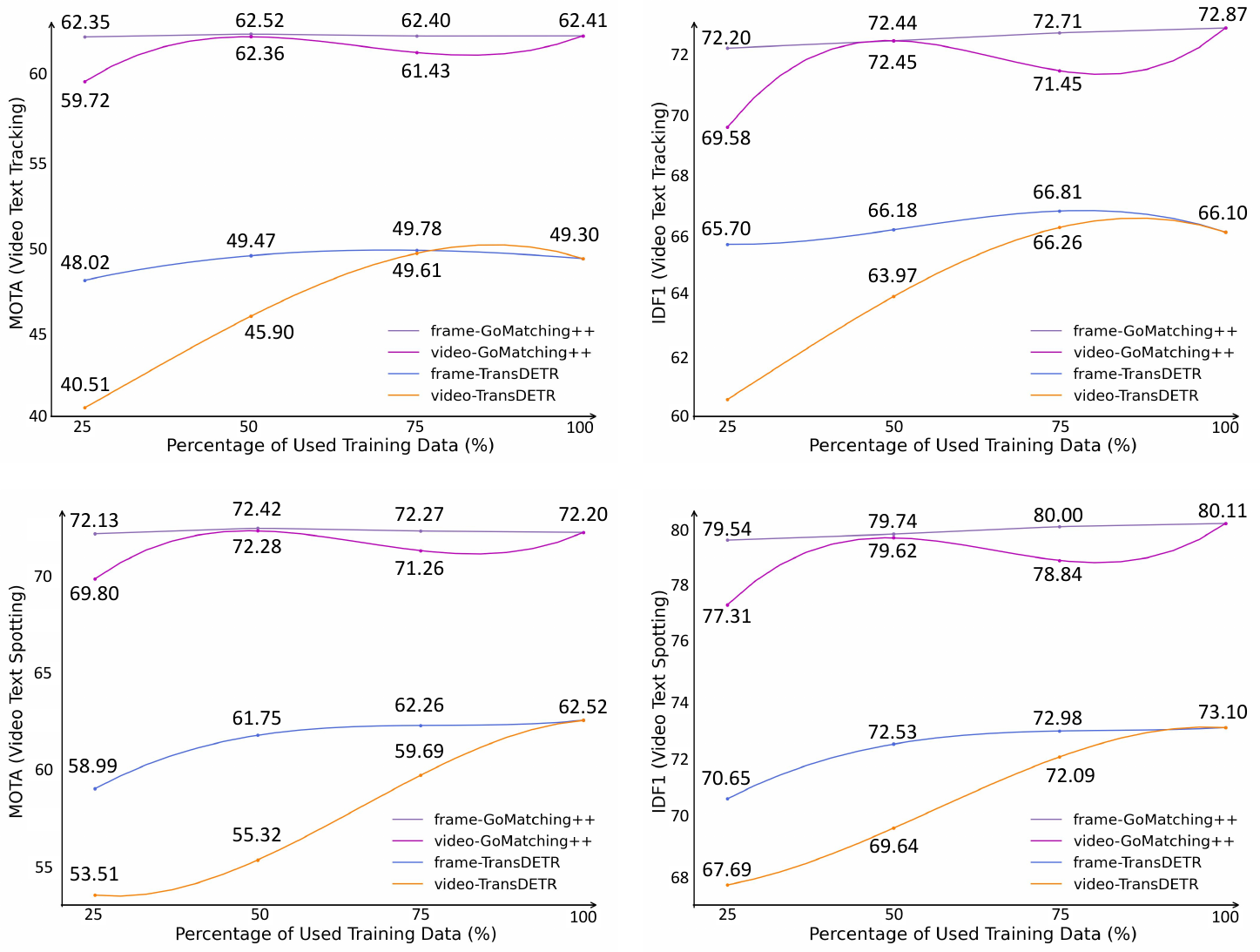}
\caption{
\textbf{Experimental Results on ICDAR15-video with different training dataset scales.} `Frame' for results from training on a specified proportion of randomly extracted frames from video clips, while `Video' for results from training on a specified proportion of randomly selected video clips.}
 \label{fig:8}
\end{figure*}

\begin{table*}[t]
\begin{minipage}{\columnwidth}
    \setlength{\tabcolsep}{7pt}
    \small 
    \centering
    \caption{\textbf{Results of using different image sizes on ICDAR15-video.} The values in parentheses mean the size of the shorter side of the input image during inference. The best results are highlighted in \textbf{bold}.}
    \begin{tabular}{ccccc}
    \hline
    Method &MOTA &MOTP &IDF1 &FPS \\
    \hline
    TransDETR(800) &60.96 &74.61 &72.80 &12.69  \\
    GoMatching(800) &68.51  &77.52  &76.59 &\textbf{14.41} \\
    GoMatching(1000) &72.04  &\textbf{78.53}  &\textbf{80.11} &10.60 \\
    GoMatching++(1000) &\textbf{72.20}  &78.52  &\textbf{80.11} &11.01 \\
    \hline
    \end{tabular}
    \label{table:8}
\end{minipage}
\hfill
\begin{minipage}{\columnwidth}
    \setlength{\tabcolsep}{6pt}
    \small 
    \centering
    \caption{\textbf{Comparison between TransDETR and GoMatching.} `T-Para.', `Mem.', and `GPU-hours' refer to the number of trainable parameters, memory usage, and GPU hours consumed during training, respectively.}
    \begin{tabular}{cccc}
    \hline
    Method & \#T-Para. (M) & \#Mem. (G) & \#GPU-hours \\
    \hline
    TransDETR &39.35  &26.7 & 304 \\
    GoMatching &32.79  &8.3 & 3 \\
    GoMatching++ &11.80  &7.3 & 3 \\
    \hline
    \end{tabular}
    \label{table:9}
\end{minipage}
\end{table*}

\subsection{Comparison with State-of-the-art Methods}
\label{exp_open_cmp}
\textbf{Results on ICDAR15-video.}
To evaluate the effectiveness of our method on oriented video text, we compare it with SOTA approaches on ICDAR15-video, as shown in Tab.~\ref{table:7}. GoMatching and GoMatching++ outperform all other methods across all evaluation metrics, with GoMatching++ achieving the best performance. By integrating a robust image text spotter with a powerful tracker, GoMatching++ improves performance by 7.87\% MOTA, 1.02\% MOTP, and 1.66\% IDF1 for tracking, and by 5.24\% MOTA, 0.74\% MOTP, and 2.55\% IDF1 for spotting. Additionally, GoMatching++ surpasses the current SOTA single-model method, TransDETR, with improvements of 13.55\% MOTA, 3.93\% MOTP, and 7.54\% IDF1 for tracking, and 11.08\% MOTA, 3.92\% MOTP, and 7.31\% IDF1 for spotting.

\noindent\quad\textbf{Results on BOVText.}
Our methods, except for English text, are easily adaptable to other video text scenarios, such as bilingual textline spotting. For BOVText, which focuses on English and Chinese textline spotting, we use DeepSolo trained on bilingual datasets and fine-tune GoMatching and GoMatching++ on BOVText. The results, shown in Tab.~\ref{table:7}, demonstrate that GoMatching++ sets a new record, outperforming previous methods significantly, with the exception of the IDF1 metric in video tracking. Notably, GoMatching++ excels in video text spotting, surpassing the previous SOTA method, CoText~\cite{wu2024bilingual}, with improvements of 42.6\% in MOTA, 6.8\% in MOTP, and 15.7\% in IDF1. This outstanding performance on BOVText highlights GoMatching++'s ability to effectively spot both Chinese and English text in videos, and its flexibility for other languages through customizable image text spotters. The gap in tracking and spotting results of prior methods further underscores that recognition ability remains the key limitation.

\noindent\quad\textbf{Results on DSText.}
We further conducted experiments on DSText, focusing on dense and small video text scenarios, with results shown in Tab.~\ref{table:7}. Notably, many top methods on the DSText leaderboard rely on model ensembles and large public datasets to boost performance~\cite{Wu2023DSText}. For instance, \textit{TencentOCR} combines detection outputs from DBNet~\cite{liao2020real} and Cascade MaskRCNN~\cite{cai2019cascade}, integrates the Parseq~\cite{bautista2022scene} text recognizer, and improves end-to-end tracking with ByteTrack~\cite{zhang2022bytetrack}. Similarly, \textit{DA} uses Mask R-CNN~\cite{he2017mask} and DBNet for text detection, BotSORT~\cite{aharon2022bot} to replace the tracker in VideoTextSCM~\cite{gao2021video}, and Parseq for recognition.
As a single model, GoMatching++ slightly trails behind ensemble-based methods in video text tracking. However, this performance gap can be reduced with stronger detectors. On video text spotting task, our method shows strong capabilities, ranking first in MOTA (23.23\%), second in MOTP (80.43\%), and third in IDF1 (46.24\%). Compared to the best single-model approach, GoMatching++ achieves significant improvements: +9.04\% in MOTA and +2.83\% in IDF1 for video text tracking, and +45.86\% in MOTA and +19.81\% in IDF1 for video text spotting.

Visual results on three open-source datasets, shown in Fig.~\ref{fig:7}, demonstrate GoMatching++'s strong spotting performance on straight text across various video scenarios.

\begin{figure*}[t]
\centering
  \includegraphics[width=1.0\textwidth]{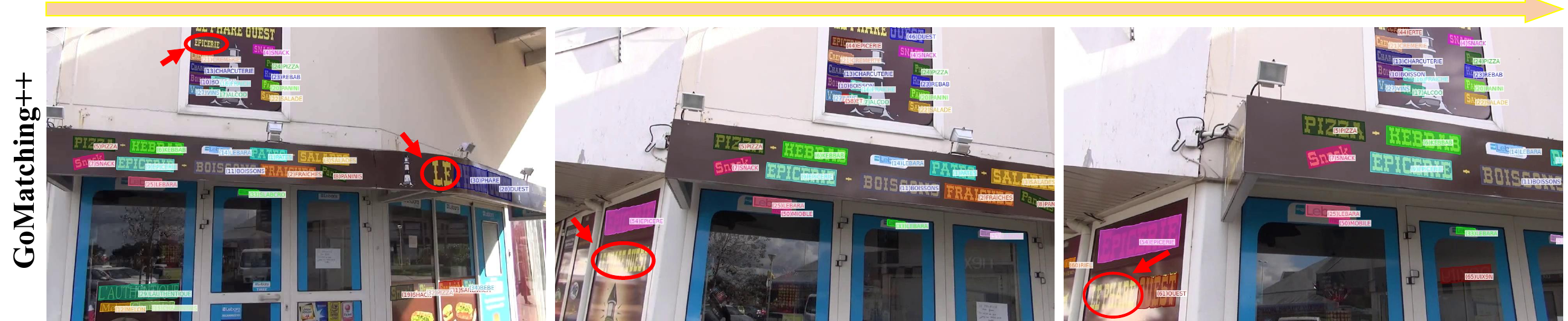}
  \caption{
  \textbf{Visualization of failure cases.} Failure cases are highlighted with ellipses. GoMatching++ may miss detections due to challenging conditions, such as tiny text and reflections.}
 \label{fig:9} 
\end{figure*}

\begin{figure*}[t]
\centering
\resizebox{\textwidth}{!}{
\begin{tabular}{cccc}
  \includegraphics[width=0.25\textwidth]{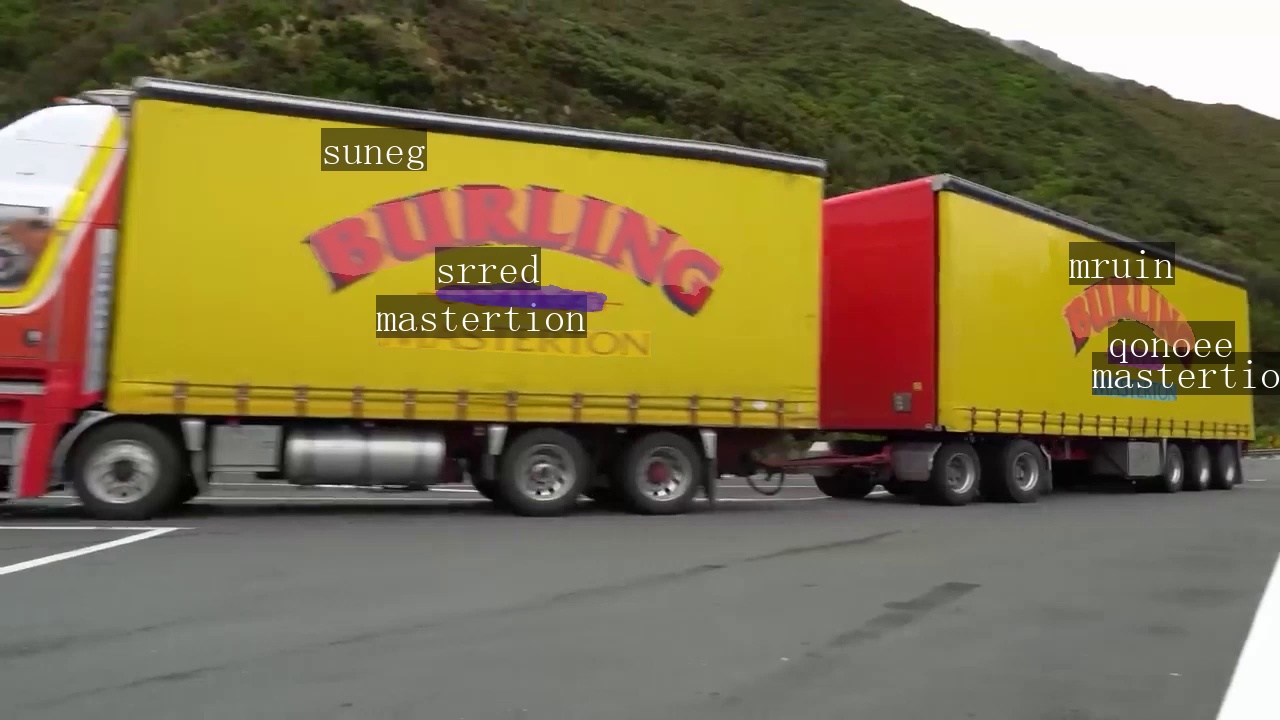} 
  &\includegraphics[width=0.25\textwidth]{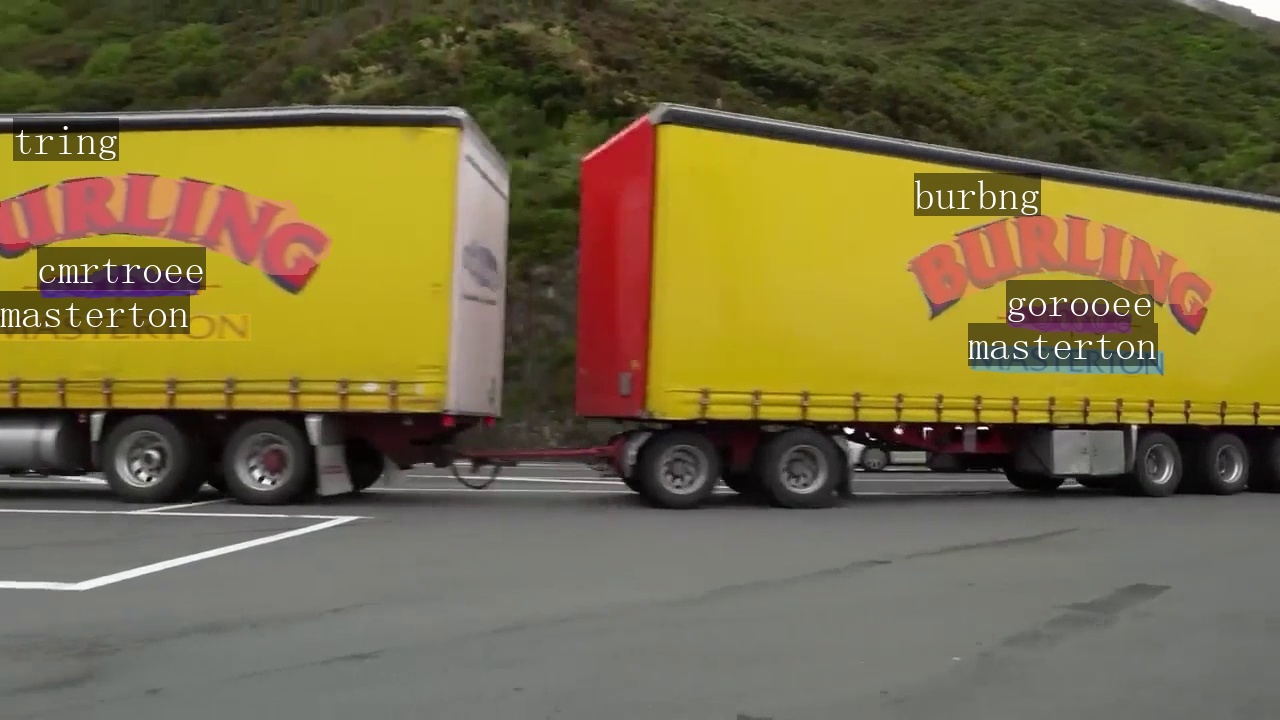}
  &\includegraphics[width=0.25\textwidth]{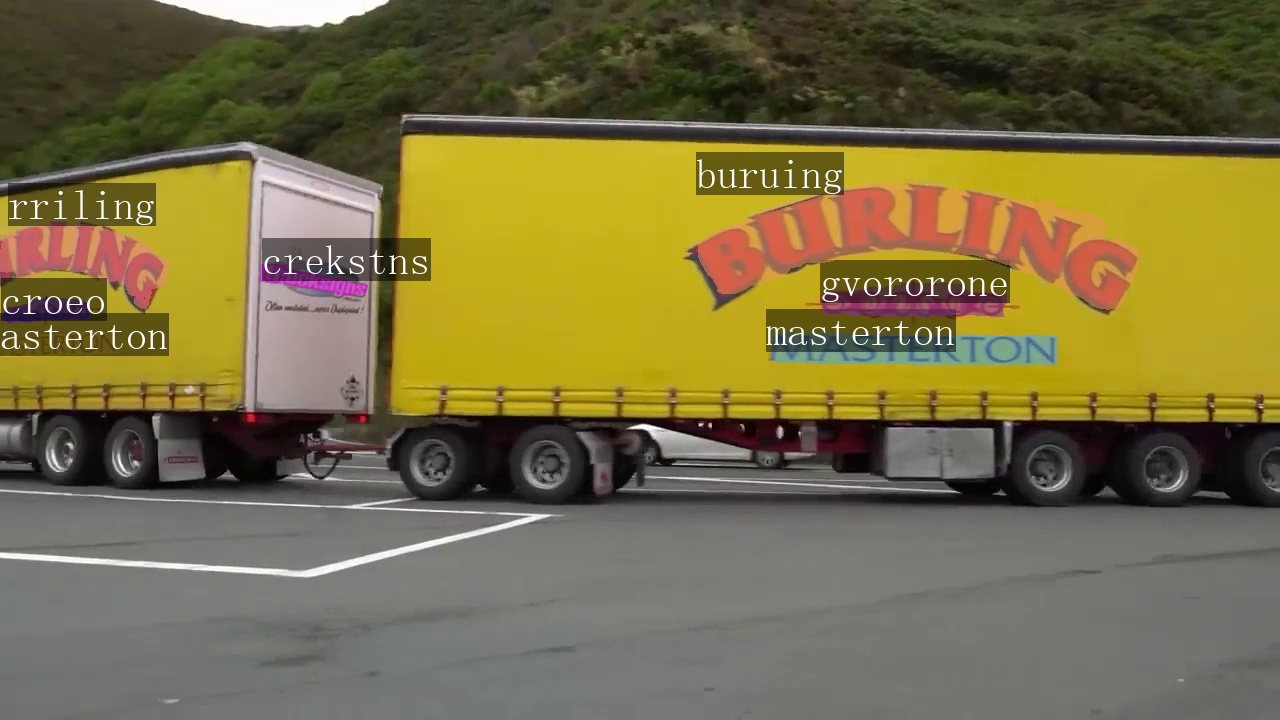}
  &\includegraphics[width=0.25\textwidth]{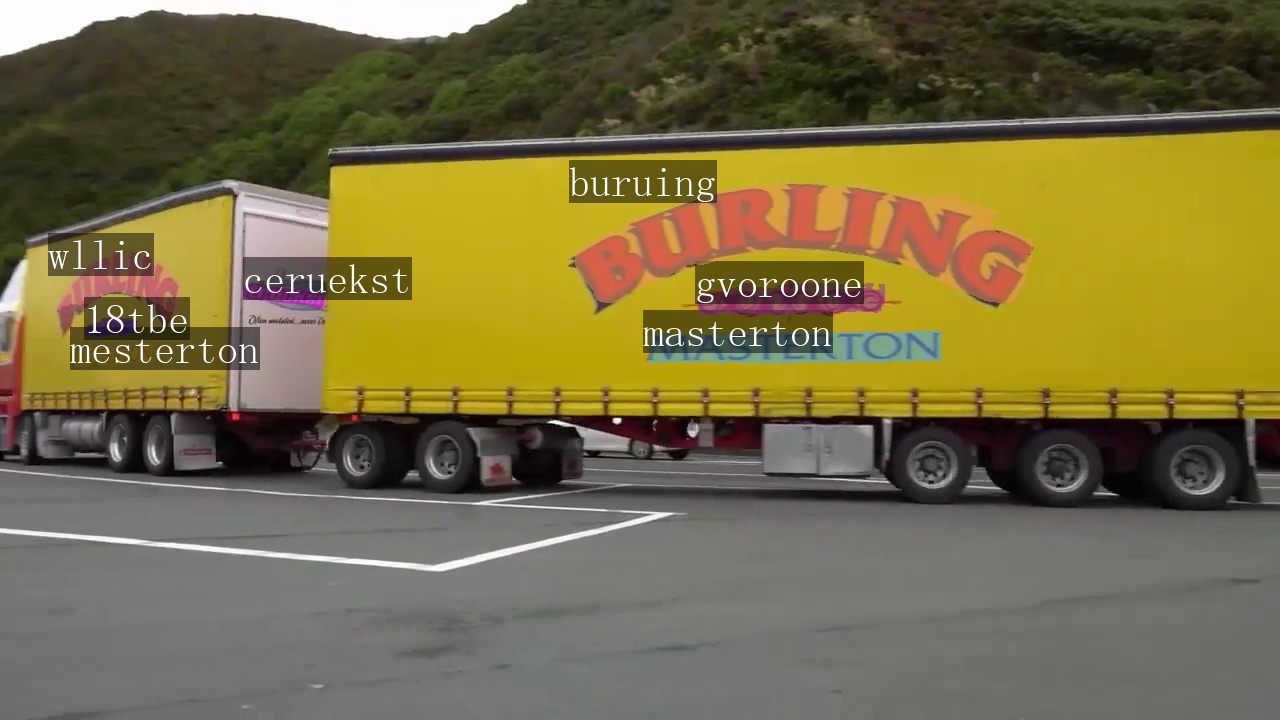}
  \\ 
  \includegraphics[width=0.25\textwidth]{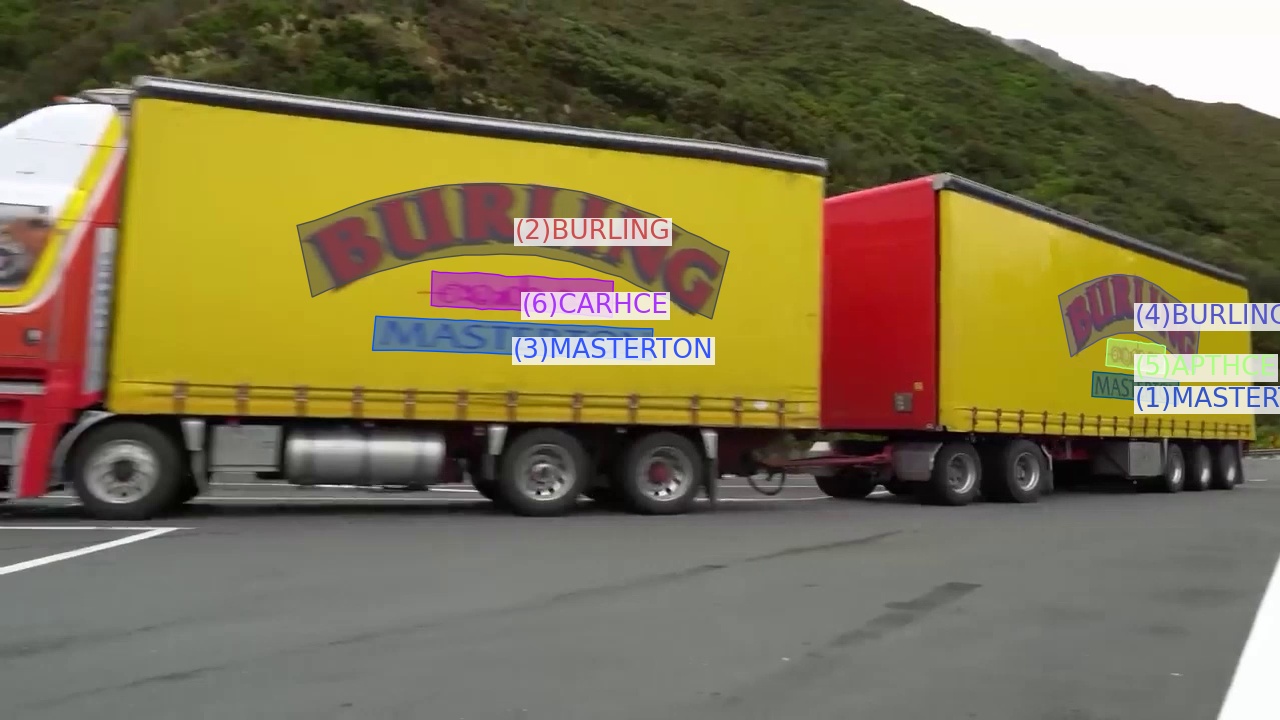} 
  &\includegraphics[width=0.25\textwidth]{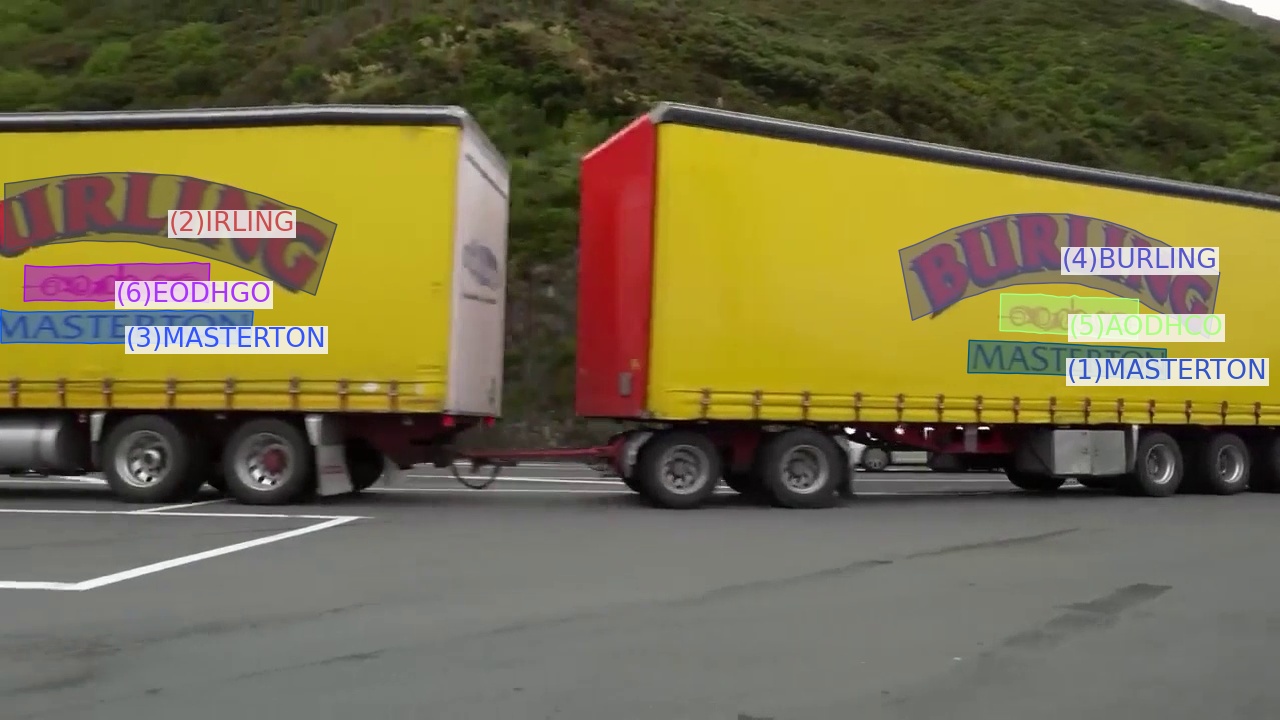}
  &\includegraphics[width=0.25\textwidth]{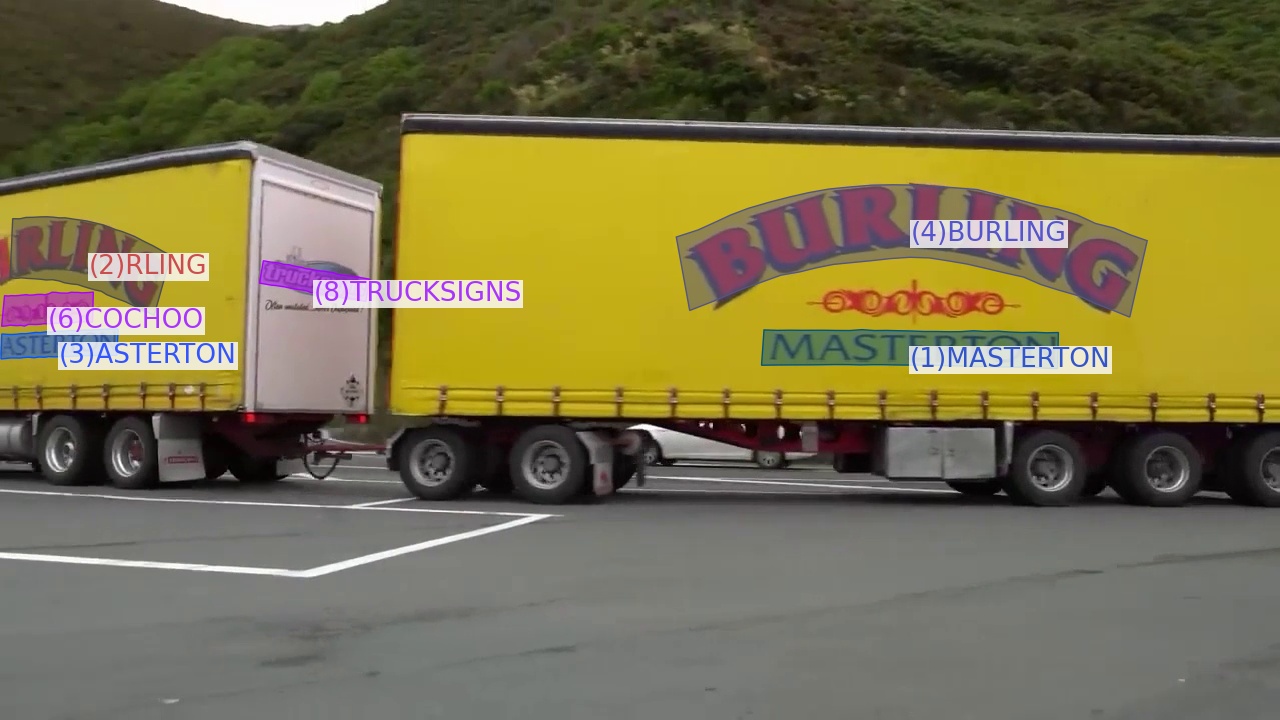}
  &\includegraphics[width=0.25\textwidth]{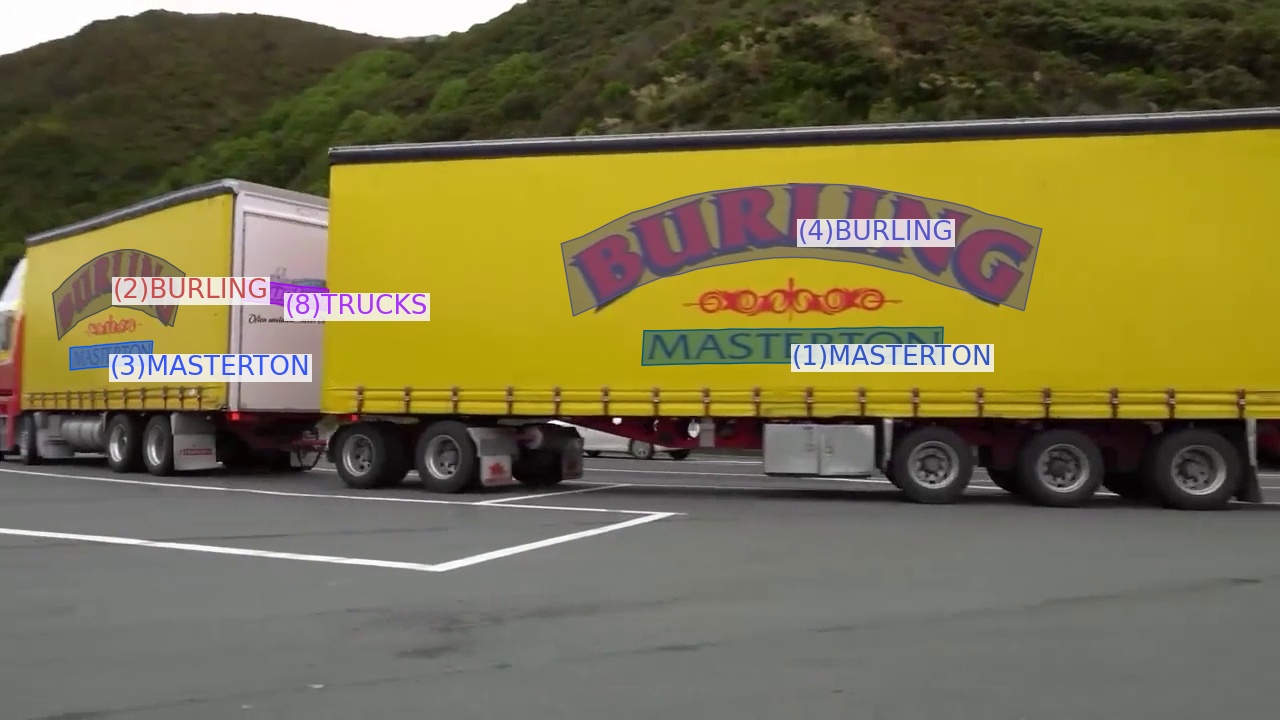}
  \\
  \end{tabular}}
\caption{
\textbf{Visual spotting results on ArTVideo.} Top to bottom: results of TransDETR-mask and GoMatching++. Text instances associated with the same trajectories are assigned the same colors.}
 \label{fig:10}
\end{figure*}

\noindent\quad\textbf{More Comparisons between TransDETR and GoMatching++.}
We further present additional comparisons between our methods and TransDETR on ICDAR15-video. In the first row of Tab.~\ref{table:8}, the input image's shorter side is set to 800, matching the default configuration of TransDETR. As shown in Tab.~\ref{table:8}, GoMatching consistently outperforms TransDETR across all settings. Notably, it exceeds TransDETR in both inference speed and spotting accuracy under its default configuration. When the image size increases (\textit{e.g.}, to 1,000), GoMatching offers superior spotting performance, albeit with a slight reduction in inference speed. With the lighter LST-Matcher design, GoMatching++ achieves better performance and faster inference, highlighting its efficiency and effectiveness.
Additionally, we compare the number of trainable parameters and training costs in Tab.~\ref{table:9}. GoMatching++ requires significantly less training budget than TransDETR due to its simpler architecture, with a 70\% reduction in trainable parameters, 19.4GB less memory usage, and 301 fewer GPU-hours. This makes GoMatching++ a more efficient and competitive baseline for video text spotting.

Moreover, we conduct experiments on ICDAR15-video to assess the data efficiency of different methods, with results shown in Fig. \ref{fig:8}. Four scale settings (25\%, 50\%, 75\%, 100\%) are applied to both `frame' and `video' training. In `frame' training, a specific proportion of frames from video clips is used, while in `video' training, a proportion of entire video clips is selected. Due to minimal variation between adjacent frames, using just 25\% of the frames does not notably affect tracking performance. However, for the recognition task (spotting), TransDETR's end-to-end training struggles with insufficient training of the recognition head, causing a significant drop in spotting performance. This issue is exacerbated when only 25\% of video clips are used (6 clips), leading to a 9.01\% drop in MOTA. In contrast, GoMatching++ exhibits better data efficiency, showing minimal performance loss (2.4\% decrease in MOTA) even with just 6 video clips.

\subsection{Failure Cases Analysis}\label{exp_open_fail}
We present additional visualization results of GoMatching++ in Fig.~\ref{fig:9}, including failure cases. Notably, GoMatching++ may struggle with extreme conditions, such as tiny text and reflections. These challenges can be addressed by using a more powerful text spotter with a robust backbone and training on larger, more diverse datasets.

\begin{table*}[t]
    \centering
    \small
    \setlength{\tabcolsep}{15pt}
    \begin{tabular}{c|ccc|ccc}
    \hline
         \multirow{2}{*}{Method} & MOTA ($\uparrow$) &MOTP ($\uparrow$) &IDF1 ($\uparrow$) &MOTA ($\uparrow$) &MOTP ($\uparrow$) &IDF1 ($\uparrow$) \\ \cline{2-7}
         & \multicolumn{3}{c|}{Video Text Tracking}  &\multicolumn{3}{c}{Video Text Spotting} \\
         \hline
         TransDETR-mask$^\bigstar$ &65.5 &82.3 &74.6 &21.1 &86.2 &59.3 \\
         GoMatching &67.9 &82.4 &76.5 &75.6 &83.5 &81.5 \\
         GoMatching++ &68.2 &82.5 &76.4 &75.7 &83.5 &82.3 \\
         \hline
         & \multicolumn{3}{c|}{Video Curved Text Tracking}  &\multicolumn{3}{c}{Video Curved Text Spotting}  \\ \cline{2-7}
         \hline
         TransDETR-mask$^\bigstar$ &42.1 &82.7 &71.6 &-25.5 &76.4 &47.0 \\
         GoMatching &65.8 &76.4 &76.4 &71.0 &78.0 &79.1 \\
         GoMatching++ &67.1 &76.9 &76.9 &73.3 &78.0 &80.0 \\
         \hline
    \end{tabular}
    \caption{
    \textbf{Intra-dataset evaluation results.} Models are trained on the ArTVideo training set and evaluated on the ArTVideo test set. `$\bigstar$' indicates that the structure of TransDETR is modified to utilize polygon and mask annotations in ArTVideo.
    }
    \label{table:10}
\end{table*}

\section{Experiments on ArTVideo}\label{exp_art}
\subsection{Implementation Details}\label{exp_art_imp}
We train GoMatching and GoMatching++ under the same settings outlined in Sec.~\ref{exp_open_det}, using a 1280-pixel input image size during inference. The original TransDETR model, designed for quadrilateral bounding boxes, does not fully leverage the fine-grained annotations (polygons and masks) in ArTVideo. To address this, we enhance TransDETR with an additional mask branch, inspired by Mask R-CNN \cite{he2017mask}, creating a new variant called TransDETR-mask. This modification enables the model to utilize mask annotations from ArTVideo, improving the accuracy of text contour predictions. Apart from this change, all other settings follow the default configuration in the open-source implementation.

\begin{table*}[t]
    \centering
    \small
    \setlength{\tabcolsep}{7pt}
    \begin{tabular}{c|ccc|ccc}
    \hline
    \multicolumn{7}{c}{\textbf{Test on ArTVideo}} \\
    \hline
        \multirow{2}{*}{Method (Training Set)} &MOTA ($\uparrow$) &MOTP ($\uparrow$) &IDF1 ($\uparrow$) &MOTA ($\uparrow$) &MOTP ($\uparrow$) &IDF1 ($\uparrow$) \\ \cline{2-7}
         & \multicolumn{3}{c|}{Video Text Tracking}  &\multicolumn{3}{c}{Video Text Spotting} \\
         \hline
         TransDETR (ICDAR15-video)* &54.2 &67.9 &70.4 &2.8 &69.7 &49.3 \\
         TransDETR-mask (ArTVideo)$^\bigstar$ &65.5(\darkgreen{+11.3}) &82.3(\darkgreen{+14.4}) &74.6(\darkgreen{+4.2}) &21.1(\darkgreen{+18.3}) &86.2(\darkgreen{+16.8}) &59.3(\darkgreen{+10.0}) \\
         \rowcolor{gray!20} 
         GoMatching (ICDAR15-video)&67.1 &82.1 &76.1 &70.1 &83.5 &79.8 \\
         \rowcolor{gray!20}
         GoMatching (ArTVideo)&67.9(\darkgreen{+0.8}) &82.4(\darkgreen{+0.3}) &76.5(\darkgreen{+0.4}) &75.6(\darkgreen{+5.5}) &83.5(\darkgreen{+0.0}) &81.5(\darkgreen{+1.7}) \\
         \rowcolor{gray!40}
         GoMatching++  (ICDAR15-video)&67.1 &82.1 &76.4 &70.1 &83.5 &80.5 \\
         \rowcolor{gray!40}
         GoMatching++  (ArTVideo)&68.2(\darkgreen{+1.1}) &82.5(\darkgreen{+0.4}) &76.4(\darkgreen{+0.0}) &75.7(\darkgreen{+5.6}) &83.5(\darkgreen{+0.0}) &82.3(\darkgreen{+1.8}) \\
         \hline
         & \multicolumn{3}{c|}{Video Curved Text Tracking}  &\multicolumn{3}{c}{Video Curved Text Spotting}  \\ 
         \hline
         TransDETR (ICDAR15-video)* &4.4 &60.5 &50.2 &-66.7 &61.9 &26.9 \\
         TransDETR-mask (ArTVideo)$^\bigstar$ &42.1(\darkgreen{+37.7}) &82.7(\darkgreen{+22.2}) &71.6(\darkgreen{+21.4}) &-25.5(\darkgreen{+41.2}) &76.4(\darkgreen{+14.5}) &47.0(\darkgreen{+20.1}) \\
         \rowcolor{gray!20}
         GoMatching (ICDAR15-video)&61.1 &76.1 &74.2 &60.5 &78.0 &75.4 \\
         \rowcolor{gray!20}
         GoMatching (ArTVideo)&65.8(\darkgreen{+4.7}) &76.4(\darkgreen{+0.3}) &76.4(\darkgreen{+2.2}) &71.0(\darkgreen{+10.5}) &78.0(\darkgreen{+0.0}) &79.1(\darkgreen{+3.7}) \\
         \rowcolor{gray!40}
         GoMatching++  (ICDAR15-video)&61.2 &76.6 &75.6 &60.6 &78.0 &77.0 \\
         \rowcolor{gray!40}
         GoMatching++  (ArTVideo)&67.1(\darkgreen{+5.9}) &76.9(\darkgreen{+0.0}) &76.9(\darkgreen{+1.3}) &73.3(\darkgreen{+12.7}) &78.0(\darkgreen{+0.0}) &80.0(\darkgreen{+3.0}) \\
         \hline
         \hline
    \multicolumn{7}{c}{\textbf{Test on ICDAR15-video}} \\
    \hline
        \multirow{2}{*}{} & \multicolumn{3}{c|}{Video Text Tracking}  &\multicolumn{3}{c}{Video Text Spotting}  \\ \cline{2-7}
         \hline
         TransDETR (ICDAR15-video)* &48.86 &74.01 &65.33 &60.96 &74.61 &72.80 \\
         TransDETR-mask (ArTVideo)$^\bigstar$ &49.31(\darkgreen{+0.45}) &74.41(\darkgreen{+0.40}) &62.86(\darkred{-2.47}) &60.97(\darkgreen{+0.01}) &74.87(\darkgreen{+0.26}) &70.49(\darkred{-2.31}) \\
         \rowcolor{gray!20}
         GoMatching (ICDAR15-video)&62.24 &77.94 &72.81 &72.04 &78.53 &80.11 \\
         \rowcolor{gray!20}
         GoMatching (ArTVideo)&62.06(\darkred{-0.18}) &77.59(\darkred{-0.35}) &72.74(\darkred{-0.07}) &68.67(\darkred{-3.37}) &78.45(\darkred{-0.08}) &77.67(\darkred{-2.44}) \\
         \rowcolor{gray!40}
         GoMatching++  (ICDAR15-video)&62.41 &77.94 &72.87 &72.20 &78.52 &80.11 \\
         \rowcolor{gray!40}
         GoMatching++  (ArTVideo)&63.02(\darkgreen{+0.61}) &77.57(\darkred{-0.37}) &73.79(\darkgreen{+0.92}) &69.30(\darkred{-2.90}) &78.47(\darkred{-0.05}) &78.77(\darkred{-1.34}) \\
         \hline
    \end{tabular}
    \caption{
    \textbf{Inter-dataset evaluation results.} `*': Model evaluated using the officially released version. `$\bigstar$': TransDETR structure modified to support polygon and mask annotations in ArTVideo.}
    \label{table:11}
\end{table*}

\subsection{Intra-Dataset Evaluation}\label{exp_art_intra}
To evaluate the effectiveness of our methods for curved text, we compare GoMatching and GoMatching++ with TransDETR-mask on the ArTVideo dataset. As shown in Tab.~\ref{table:10}, both GoMatching and GoMatching++ outperform TransDETR-mask across all evaluation tasks in curved-text setting. The performance gap is particularly notable when adding a recognition task for end-to-end spotting or focusing solely on curved text. GoMatching++ achieves a 98.8\% improvement in MOTA and a 33.0\% improvement in IDF1 over TransDETR-mask. These results highlight the limitations of prior SOTA methods, which struggle with recognition and adaptability due to the inherent design of their frameworks. Furthermore, the performance boost of GoMatching++ over GoMatching emphasizes the effectiveness of its optimized design. Visualization results in Fig.~\ref{fig:10} demonstrate a clear improvement in text recognition by GoMatching++ compared to TransDETR-mask.

\subsection{Inter-Datasets Evaluation}\label{exp_art_inter}
In this subsection, we explore the transferability and generalization of models across datasets through inter-dataset evaluations between the ICDAR15-video and ArTVideo datasets. As shown in Tab.~\ref{table:11}, the lack of curved text in ICDAR15-video results in significantly lower performance when models trained on ICDAR15-video are tested on ArTVideo, especially for tasks involving curved text. For example, TransDETR, trained on ICDAR15-video, experiences a 37.7\% drop in MOTA and 21.4\% in IDF1 for video curved text tracking, and a 41.2\% drop in MOTA and 20.1\% in IDF1 for video curved text spotting compared to models trained on ArTVideo. GoMatching and GoMatching++ perform better, although they still experience a 10\% drop in MOTA due to the lack of curved text in the training data. In contrast, models trained on ArTVideo and tested on ICDAR15-video perform similarly to those trained directly on ICDAR15-video, with performance decreases not exceeding 4\%, and some metrics even showing improvements. These results demonstrate GoMatching++'s superiority over prior SOTA methods and highlight the diversity of the proposed ArTVideo dataset and its potential to enhance model robustness across various video text spotting tasks.

\section{Conclusion}\label{conclusion}
This paper introduces GoMatching and its enhanced version, GoMatching++, as effective video text spotting baselines with a concise and efficient framework. GoMatching++ contributes three key innovations: (1) transforming an off-the-shelf image text spotter into a video specialist with a lightweight, trainable tracker; (2) bridging the image-video domain gap with a rescoring mechanism; and (3) improving video text handling via the LST-Matcher with minimal training data. We also establish ArTVideo, a new benchmark for arbitrary-shaped text in video, with a comprehensive analysis of its data sources, annotations, and key characteristics. Extensive experiments show that GoMatching++ achieves superior performance and efficiency in both parameter and data usage. We believe our methods and the ArTVideo benchmark will drive future advancements in video text spotting, particularly for complex, curved text in real-world scenarios.

\section*{Acknowledgements}
This work was supported by WHU-Kingsoft Joint Lab. The numerical calculations in this paper have been done on the supercomputing system in the Supercomputing Center of Wuhan University.

\bibliographystyle{IEEEtran}
\bibliography{ref}

\vfill

\end{document}